\documentclass[10pt,twocolumn,letterpaper]{article}

\usepackage{cvpr}              

\usepackage[table]{xcolor}

\usepackage{amsmath,amsfonts,bm}

\def\eqref#1{equation~\ref{#1}}

\def\vx{{\mathbf{x}}}

\DeclareMathAlphabet{\mathsfit}{\encodingdefault}{\sfdefault}{m}{sl}
\SetMathAlphabet{\mathsfit}{bold}{\encodingdefault}{\sfdefault}{bx}{n}

\DeclareMathOperator*{\argmax}{arg\,max}
\DeclareMathOperator*{\argmin}{arg\,min}

\newcommand{\bM}{\mathbf{M}}
\newcommand{\bQ}{\mathbf{Q}}

\newcommand{\bc}{\mathbf{c}}

\newcommand{\bv}{\mathbf{v}}
\newcommand{\bw}{\mathbf{w}}

\newcommand{\bx}{\mathbf{x}}

\newcommand{\bz}{\mathbf{z}}

\newcommand{\bbE}{\mathbb{E}}

\newcommand{\bbR}{\mathbb{R}}

\newcommand\egaldef{\stackrel{\mbox{\rm\tiny def}}{=}}

\newcommand{\bff}{\mathbf{f}}

\usepackage{comment}

\usepackage{lipsum}
\usepackage{algorithm}
\usepackage{algorithmic}
\usepackage{listings}

\usepackage{mathtools}
\usepackage{physics}

\usepackage{amssymb}
\usepackage{amsmath}

\newcommand{\tao}[1]{~\textcolor{red}{tao:\ #1}}

\usepackage{stackengine,scalerel}

\definecolor{lightergray}{RGB}{220,220,220} 
\definecolor{mygray}{gray}{0.95}

\usepackage{pifont}
\newcommand{\cmark}{\ding{51}}%
\newcommand{\xmark}{\ding{55}}%

\usepackage{xcolor}

\definecolor{cvprblue}{rgb}{0.21,0.49,0.74}
\usepackage[pagebackref,breaklinks,colorlinks,citecolor=cvprblue]{hyperref}

\def\paperID{265} 
\def\confName{CVPR}
\def\confYear{2024}

\title{
Guided Diffusion from Self-Supervised Diffusion Features
}

\author
{
Vincent Tao Hu$^{1,2}\thanks{Part of this work was done while at University of Amsterdam.}$,
Yunlu Chen$^3$,
Mathilde Caron$^4$,\\
Yuki M. Asano$^2$,
Cees G. M. Snoek$^2$,
Björn Ommer$^1$\\
$^1$CompVis Group, LMU Munich, $^2$University of Amsterdam, $^3$CMU, $^4$Google Research \\
}

\begin{document}
\maketitle
\begin{abstract}
Guidance serves as a key concept in diffusion models, yet its effectiveness is often limited by the need for extra data annotation or classifier pretraining. That is why guidance was harnessed from self-supervised learning backbones, like DINO. However, recent studies have revealed that the feature representation derived from diffusion model itself is discriminative for numerous downstream tasks as well, which prompts us to propose a framework to extract guidance from, and specifically for, diffusion models. 
Our research has yielded several significant contributions. Firstly, the guidance signals from diffusion models are on par with those from class-conditioned diffusion models. Secondly, feature regularization, when based on the Sinkhorn-Knopp algorithm, can further enhance feature discriminability in comparison to unconditional diffusion models. Thirdly, we have constructed an online training approach that can concurrently derive guidance from diffusion models for diffusion models. Lastly, we have extended the application of diffusion models along the constant velocity path of ODE to achieve a more favorable balance between sampling steps and fidelity. The performance of our methods has been outstanding, outperforming related baseline comparisons in large-resolution datasets, such as ImageNet256, ImageNet256-100 and LSUN-Churches. Our code will be released.
\end{abstract}

%

%
%
\vspace{-18pt}
\section{Introduction}

\begin{figure}
    \centering
    \includegraphics[width=0.5\textwidth, trim={0 17.8cm 29.5cm 0},clip]{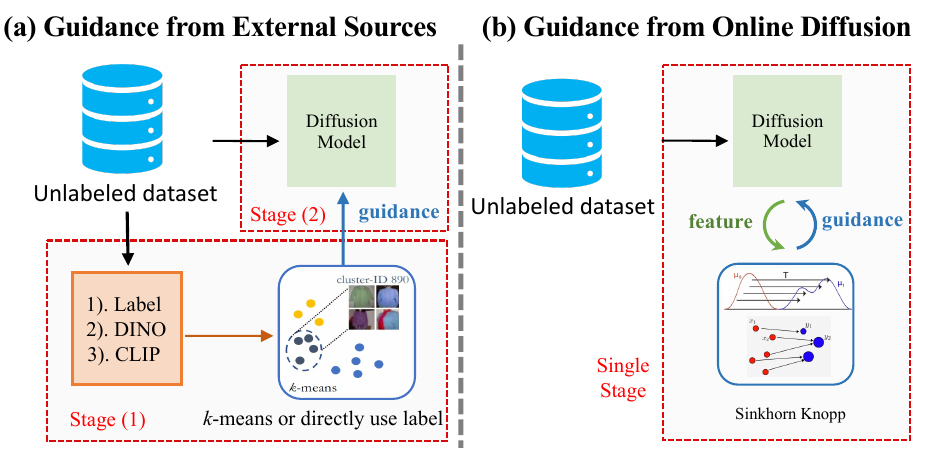}
     \vspace{-17pt}
    \caption{Previous works derive the external guidance signal from  burdensome label annotation, or self-supervised representations to improve fidelity in a complex two-stage manner. Our work, however, boosts performance by using the diffusion model itself to achieve online self-guidance in a single stage. }
    \label{fig:teaser}
\end{figure}

Diffusion models have shown significant advancements in computer vision, including image processing~\cite{rombach2022high_latentdiffusion_ldm}, video analysis~\cite{ho2022video}, point cloud processing~\cite{wu2023fast}, and human pose estimation~\cite{gong2023diffpose}. Typically, these applications rely on guidance signals to enhance controllability based on user preferences, resulting in improved fidelity~\cite{dhariwal2021diffusion_beat,saharia2022photorealistic_imagen}. However, these applications often heavily depend on labor-intensive annotations, such as image-text pairs~\cite{radford2021learning_clip} or image-label pairs~\cite{imagenet}. Annotating the data is time-consuming,  error-prone and may inject bias. In our work, we aim to address this guidance requirement by eliminating the need for data annotations. 

A few studies~\cite{sgdm, bao2022conditional, harvey2023visual_cot} have explored the possibility of harnessing self-supervised representations to provide annotation-free training for diffusion models. 
Concurrently, the features extracted by diffusion models themselves have found utility across diverse representation learning~\cite{yang2023diffusion_representation_freelunch} and image correspondence tasks~\cite{tang2023emergent_correspondence_from_dm,mou2023dragondiffusion}, which illustrates the significant potential for leveraging diffusion models in discriminative tasks.
%
On the other hand, diffusion models have evolved into a more versatile form, encompassing both the Stochastic Differential Equation (SDE) approach, such as Variance-Preserving SDE~\cite{ho2020denoising} and {Variance-Exploding} SDE~\cite{song2019generative}, as well as the Ordinary Differential Equation (ODE) approach~\cite{liu2022flow,lipman2022flow,albergo2023stochastic}. Notably, ODE-based models tend to strike a more favorable balance between sampling efficiency and quality compared to SDE-based counterparts, as demonstrated in~\cite{karras2022elucidating}. Given that self-guidance and generalized diffusion models offer two crucial components for advancing generative learning, the natural progression is to explore the integration of self-guidance within the framework of generalized diffusion models.

In this work, we introduce a novel framework that harnesses the inherent guidance capabilities of the diffusion model itself, as illustrated in Figure~\ref{fig:teaser}. Our contributions encompass three key aspects: Firstly, we demonstrate the feasibility of directly extracting guidance signals from diffusion models themselves, eliminating the reliance on self-supervised learned backbones (e.g., DINO~\cite{dino}). Secondly, we introduce an online optimal-transport-based algorithm to extract guidance signals from diffusion models, yielding substantial performance improvements for these models.
%
Our approach outperforms unconditional generation by leveraging the same amount of \textit{unlabelled} data and achieves results that are comparable to both clustering-based guidance signals and class-conditioned guidance. Finally, our method not only generates images based on user query inputs but also successfully achieves an effective equilibrium between sampling speed and fidelity.

\section{Related works}

\paragraph{Exploiting features in diffusion models.} 

Diffusion Models~\cite{sohl2015deep,ho2020denoising,song2021scorebased_sde} have demonstrated their versatility across various domains~\cite{rombach2022high_latentdiffusion_ldm,saharia2022photorealistic_imagen}. Recently, it has been shown that diffusion models can be beneficial for discriminative tasks such as classification~\cite{mukhopadhyay2023diffusion_classification}, semi-supervised learning~\cite{you2023diffusion_semi}, and more specific downstream tasks like human pose estimation~\cite{gong2023diffpose,qiu2023learning,holmquist2022diffpose} and representation learning~\cite{yang2023diffusion_representation_freelunch}.
From another perspective, the intermediate features of these models have been exploited in various computer vision applications, such as image correspondence~\cite{tang2023emergent_correspondence_from_dm,mou2023dragondiffusion}, even cross-modality correspondences~\cite{wang2023freereg}, scene geometry, support relations, shadows and depth~\cite{zhan2023does_dm_3d_freelunch}, open-vocabulary object segmentation~\cite{xu2023open,li2023openvoc_seg_dm_freelunch}, and few-shot segmentation~\cite{fss_diff}. In this paper, we take a step further to explore the possibility of utilizing features from diffusion models to further bootstrap its self-guidance.


\paragraph{Guidance in diffusion models.}  Diffusion models excel in controllable generative modeling, owing to their capacity of guidance for enhanced controllability~\cite{rombach2022high_latentdiffusion_ldm} and fidelity~\cite{dhariwal2021diffusion_beat}.  To leverage the performance of pretrained diffusion models and utilize the mathematical principles underlying the score in diffusion models, Dhariwal and Nichol~\cite{dhariwal2021diffusion_beat} suggest training an additional classifier on noisy data to guide unconditional diffusion models. Complementing this, Ho \etal~\cite{ho2021classifier} propose classifier-free guidance, which involves random dropout of the guidance signal to integrate the capabilities of both unconditional and conditional sampling within a singular network. Nevertheless, these approaches still require data annotation during training.

Recent efforts~\cite{sgdm, bao2022conditional} aim to eliminate the need for data annotation, capitalizing on advancements in self-supervised learning~\cite{caron2020unsupervised_swap,asano2019selflabelling,dino}. They utilize external self-supervised learned backbones to extract guidance signals for image, box, and pixel-level guidance, achieving high-fidelity generation without relying on data annotation. Several existing works such as Slot-diffusion~\cite{wu2023slotdiffusion,jiang2023object} also address this problem through the lens of object-centric learning, often requiring extra self-supervised learning encoders, particularly in complex scenarios. Unlike all of them, our work eliminates the reliance on external self-supervised models. Instead, we exploit guidance from the diffusion model itself. This is achieved through an optimal-transport assignment during training, enabling the model to cultivate its own guidance mechanism.

\begin{figure*}
    \centering
    \includegraphics[width=0.93\textwidth]{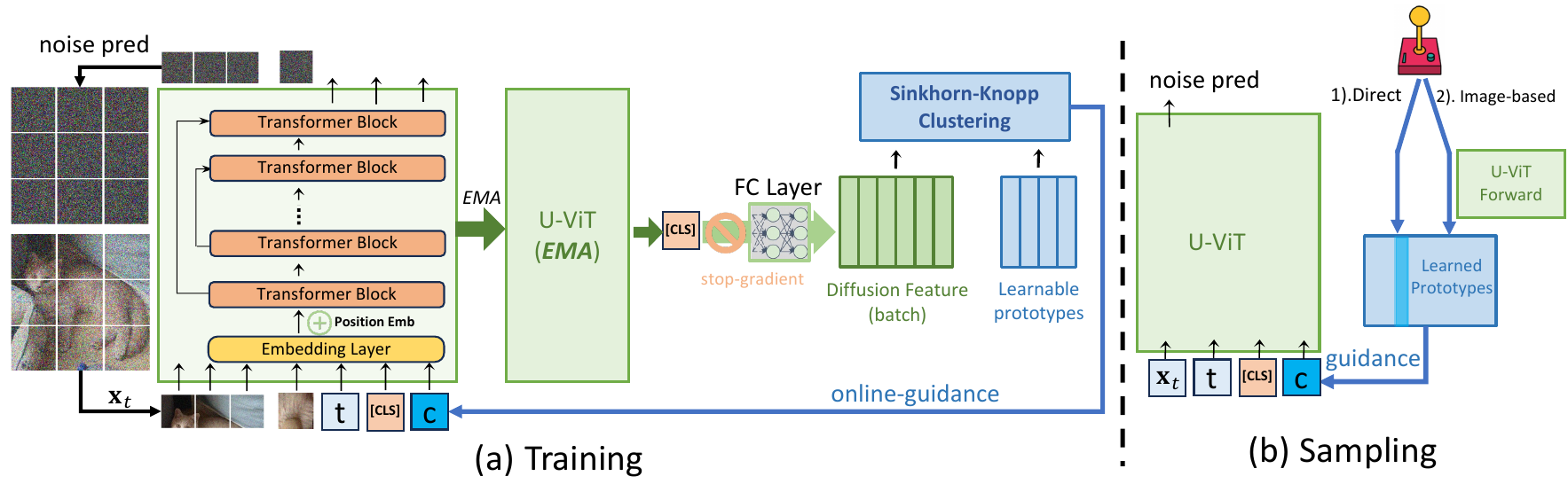}
    \vspace{-10pt}
    \caption{\textbf{Our Framework.} Our method leverages the inherent guidance capabilities of diffusion models during training by incorporating the optimal-transport loss. In the sampling phase, we can condition the generation on either the learned prototype or by an exemplar image. }
    \label{fig:main}
\end{figure*}

\vspace{-3pt}
\section{Preliminaries} 



We design our online guidance based on the generalized diffusion framework, chosen for its flexibility, simplicity, and efficient sampling capabilities. We will illustrate its basics through both stochastic and deterministic perspectives.

\paragraph{Stochastic diffusion probabilistic models.}
Given samples from an underlying data distribution $\bx_0 \sim p_0$, the goal of a diffusion probabilistic model~\cite{ho2020denoising,song2019generative} is to fit the distribution and enable the generation of novel examples through sampling. 
The diffusion process gradually corrupts the sample $\bx_0$ with noise to $\bx_t$ in a continuous time schedule $t \in [0, T]$ until $p_T(\bx_T)$ becomes random noise, described as a stochastic differential equation (SDE)~\cite{song2021scorebased_sde}:
\begin{align}
    \dd{\bx} = \bff(\bx, t) \dd{t} + g(t) \dd{\bw},
\end{align}
where $\bw$ is the standard Wiener process, $\bff(\bx, t)$ is the vector-valued drift coefficient, $g(t)$ is the scalar diffusion coefficient. The reverse-time SDE diffusion process generates data starting by sampling from $p_T(\bx_T)$~\cite{song2021scorebased_sde,anderson1982reverse}:
\begin{align}
    \dd{\bx} = [\bff(\bx, t) - g^{2} \nabla_{\bx} \log p_t(\bx)] \dd{t} + g(t) \dd{\overline{\bw}},\label{eq:sgm_sde}
\end{align}
where  $\nabla_{\bx} \log p_t(\bx)$ is known as the score function of the noise-perturbed data distribution,  $\overline{\bw}$ is a reverse-time standard Wiener process,.


\paragraph{Deterministic diffusion models.}
For any diffusion process, there exists a corresponding deterministic process with the same marginal probability densities in the form of an ODE.
As an example, Song et al.~\cite{song2021scorebased_sde} introduce a deterministic Probability flow ODE, where the velocity in probability flow ODE has the form
\begin{equation}
  \label{eq:velocity-probaflow}
  v_t(\bx_t, t) = f(\bx_t, t) - \frac{g_t^2}{2} \nabla \log p_t,
\end{equation}
where $\dd{\bx} = v_t(\bx_t, t) \dd{t}$. It is shown in \cite{karras2022elucidating,lee2023minimizing} that using ODE sampler of the form Equation~\ref{eq:velocity-probaflow} 
can reduce sampling costs compared to sampling with discretization of the diffusion SDE in Equation~\ref{eq:sgm_sde}.

When provided with empirical observations of the data distribution $\bx_0 \sim p_0$ and typically Gaussian noise distribution $\bx_1 \sim p_1$, the objective within the deterministic Ordinary Differential Equation (ODE) framework is to estimate a coupling function $\pi(p_0, p_1)$ that characterizes the transition between these two distributions.
This objective could be formulated as solving an ODE:
\begin{align}
  \label{eq:ode-lagrangian}
  \mathrm{d}\bx_t = v(\bx_t, t) \mathrm{d}t,
\end{align}
%
over time $t \in [0,1]$, where the velocity $v:\bbR^d \times [0, 1] \to \bbR^d$
is set to drive the flow from $p_0$ to $p_1$.

Usually, the drift (velocity) $v_{\theta}(\bx_t, t)$ can be represented by a neural network, and the loss can be formulated in a regression manner:
\begin{equation}
  \label{eq:fm-obj}  
  \hat{\theta} = \argmin_{\theta} \bbE_{t, \bx_t} \left[||v(\bx_t, t) - v_{\theta}(\bx_t, t) ||^2_2 \right].
\end{equation}
%

where $v(\bx, t)$ can be designed in various forms.  In the following, we offer a concise overview of three commonly-used variations of $v(\bx, t)$: probability flow ODEs with the VP, VE, and constant velocity paths.

\textbf{1). VP path.} 
%
%
\begin{equation}
  \label{eq:vp-path}
  \bx_t \egaldef \alpha_t \bx_0 + (1-\alpha_t^2)^{\frac{1}{2}} \ \bx_1 , 
\end{equation}
$\text{where } \alpha_t = e^{-\frac{1}{2}\int_0^{t} \beta(s)\mathrm{d}s}.$ This Variance Preserving (VP) path aligns with the diffusion path in DDPM~\cite{ho2020denoising}.

\textbf{2). VE path.}
\begin{equation}
  \label{eq:ve-path}
  \bx_t \egaldef  \bx_0 + \alpha_t \bx_1 ,
\end{equation}
where $\alpha_t$ is an increasing function, and $\alpha_0=0$ and $\alpha_1 \gg 1$. This Variance Exploding (VE) path aligns with the diffusion path in SMLD~\cite{song2019generative}.

%
%
%

\textbf{3). Constant velocity path.}
%
\begin{equation}
    \bx_t \egaldef (1 - t)\bx_0 + t\bx_1,
\end{equation}

\noindent where $\bx_t$ is the linear interpolation between $\bx_0$ and $\bx_1$.
This means the velocity drives the flow following the direction $v_t(\bx_t, t) = \bx_1 - \bx_0$ for any $t \in [0,1]$. We have chosen this path due to its superior sampling efficiency and quality compared to SDE-based alternatives, as shown in recent studies~\cite{karras2022elucidating,lipman2022flow,rectifiedflow_iclr23}.

All these designs offer a versatile and comprehensive diffusion framework that goes beyond the traditional vanilla diffusion~\cite{sohl2015deep,ho2020denoising,song2021scorebased_sde}. Recent works have demonstrated a preference for this latent space~\cite{dao2023flow_lfm}, utilizing it in diverse applications such as image editing~\cite{hulfm}, inversion~\cite{pokle2023training_imageinversion_fm} and achieving rapid sampling in low NFE regimes~\cite{rectifiedflow_iclr23}. These discoveries serve as inspiration for us to develop online guidance within this innovative regime, a concept we will elaborate on in the following sections.

\vspace{-3pt}
\section{Method}

We explore self-guidance in diffusion models from two distinct perspectives. Firstly, through an offline approach, we extract guidance from diffusion models and subsequently employ it in guided diffusion training. Secondly, we directly investigate online guidance within diffusion models. Both of these perspectives will be elaborated upon in the following sections.

\subsection{Offline guidance from pretrained diffusion model features}

In this section, we assess pretrained diffusion models and utilize $k$-means clustering for generating embeddings that serve as guidance signals during training. To start with, let us consider the simple case of using guidance to finetune a diffusion model, of which the parameters $\theta^*$ of the velocity network $v_{\theta^*}$ is pretrained.  

We apply a classifier-free guidance technique to the velocity network $ v_{{\theta}}$. The model is optimized for the following loss function:
\begin{align}
\mathcal{L}_\text{offline} = \bbE_{t, \bx_t} \|\bx_1 - \bx_0 - v_{{\theta}}(\bx_t, t,\bc) \|^2_2 ,
\label{eq:fm-obj-constant}
\end{align}
%
%
where $\bc$ is the conditional embedding for the guidance, to be detailed in the following paragraph. In practice $\bc$ is randomly dropped out to be replaced by a learnable null embedding $\emptyset$ with a pre-defined probability of $p_\text{drop} = 15\%$, so that our model preserves the ability of unconditioned generation. This configuration effectively results in $\bv_{\vartheta}(\bx_t, t, \emptyset)$ approximating $p(\bx_1)$, indicating that a significant portion of the network's capacity is dedicated to conditional sampling (85\%) as opposed to unconditional sampling (15\%).
%

Our primary focus centers on investigating the feature from pretrained transformer-based network U-ViT~\cite{uvit} due to its clean architecture, efficiency, scalability, and strong performance, DiT~\cite{dit_peebles2022scalable} is also another suitable network which will explore in the future. To explore high-resolution image generation, we employ U-ViT in latent space, following the approach outlined in previous works~\citep{rombach2022high_latentdiffusion_ldm,dao2023flow_lfm,hulfm}. We consider feature representations from various timesteps $t$ and layer index $l$ of the U-ViT network. In the case of offline guidance, the one-hot cluster embedding $\bc$ is achieved by the $k$-means clustering on the pretrained U-ViT features $v_{\theta^*}(\bx, t, \emptyset)$. The rationale behind these choices is discussed in more detail in the ablation experiments section~(see Experiment~\ref{sec:ablation_offline}), which also motivates us to design the online guidance in the next section.

\subsection{Bootstrap online Sinkhorn-Knopp clustering}

We previously discussed offline guidance extraction from pretrained diffusion models on the U-ViT network. To enable online guidance, we integrate online clustering algorithms into the diffusion model, as shown in Figure~\ref{fig:main}.

\paragraph{Sinkhorn-Knopp clustering.}
 Offline clustering for guidance embedding is an intuitive approach, but it suffers from issues of low efficiency and dependence on external models.
Instead, we solve online clustering as an optimal transport problem using the  Sinkhorn-Knopp algorithm~\cite{cuturi2013sinkhorn} as inspired by the image-level methods in self-supervised learning~\cite{asano2019selflabelling,caron2020unsupervised_swap} but propose to have this \textit{simultaneous}  to the training of the diffusion model.

Our goal is to achieve guidance embeddings $\bc$ from the diffusion model during the training stage to boost the quality of generation by means of conditioning. 
The challenge arises when attempting to obtain usable signals for clustering as the conditional guidance within a diffusion model that is dependent on the conditioning. 

To this end, our approach involves employing a zero vector $\mathbf{0}$ into the conditional diffusion model $v^{l}_\theta(\bx_t,t, \mathbf{0})$ for the signals used to identify the clustering.
Note that the zero vector $\mathbf{0}$  needs to be discriminated from the learnable null embedding that replaces the dropped-out guidance embedding to retain of ability of high-quality unconditional generation. 
Particularly, for each image example, $v^{l}_\theta(\bx_t,t,\mathbf{0})$ undergoes a fully-connected feature prediction head $h_\phi(\cdot)$, resulting in a $d$-dimensional image feature embedding $\bz = h_\phi\left(v^{l}_\theta(\bx_t,t, \mathbf{0})\right) \in \mathbb{R}^d$, and we denote a batch of $B$ feature vectors as $\mathbf{Z}=[\mathbf{z}_{1},\dots,\mathbf{z}_{B}] \in \mathbb{R}^{B \times d}$. 
To achieve online clusters, we map the column feature vectors in $\mathbf{Z}$ to a set of $K$  learnable prototypes  $[\mathbf{m}_1,\dots,\mathbf{m}_K]=\mathbf{M} \in \mathbb{R}^{K \times d}$ with a joint probability transportation matrix 
$\mathbf{P} \in \mathbb{R}^{K\times B}$. 

In the context of the optimal transport problem,  $\mathbf{P}$ is relaxed to be an element of a transport polytope~\cite{cuturi2013sinkhorn}:
\begin{equation}
  \mathcal{U}(\mathbf{q}_\text{r}, \mathbf{q}_\text{c}) = \left \{\mathbf{P}\in\mathbb{R}_+^{K\times B} ~|~\mathbf{P} \mathbf{1} = \mathbf{q}_\text{r}, \mathbf{P}^\top \mathbf{1} = \mathbf{q}_\text{c} \right \},
\end{equation}
where  $\mathbf{1}$ denotes vectors of all ones of the appropriate dimensions. 
$\mathbf{q}_\text{r} \in \mathbb{R}^K$ and $\mathbf{q}_\text{c} \in \mathbb{R}^B$ are interpreted as the marginal probability vectors of rows and columns in $\mathbf{P}$.
Following Asano et al.~\cite{asano2019selflabelling}, we adhere to the simple heuristic that $\mathbf{P}$ evenly partitions examples into clusters to ensure diversity, \ie  $\mathbf{q}_\text{r} = \mathbf{1} / B$ and $\mathbf{q}_\text{c} = \mathbf{1} / K$. 

In the context of the optimal transport problem~\cite{cuturi2013sinkhorn}, we define the Sinkhorn-Knopp loss function as the \emph{dual-Sinkhorn divergence} which approximates the optimal transport distance:  
\begin{align}
    &\mathcal{L}_\text{SK}= \langle \mathbf{P}^*, -\mathbf{M}\mathbf{Z}^\top \rangle_F, \quad \text{where}\label{eq:assign}\\
    \mathbf{P}^* = & \argmin_{\mathbf{P} \in \mathcal{U}(\mathbf{q}_\text{r}, \mathbf{q}_\text{c})}  \langle \mathbf{P}, -\mathbf{M}\mathbf{Z}^\top \rangle_F - \frac{1}{\lambda} H(\mathbf{P}).
\end{align}
 Here $\langle \cdot, \cdot \rangle_F$ is the Frobenius inner product.  $-\mathbf{M}\mathbf{Z}^\top$ is the distance matrix between column elements in $\mathbf{M}$ and $\mathbf{Z}$. $H(\mathbf{P}^*) = -\sum_{ij} \mathbf{P}^*_{ij} \log \mathbf{P}^*_{ij}$ is the entropy term introduced by Cuturi~\cite{cuturi2013sinkhorn} with a parameter $\lambda >0$. With this term, the optimal  $\mathbf{P}^*$  can be calculated for a much cheaper cost than the original optimal transport problem 
 in the form of:
\begin{equation}
\label{eq:qstar}
  \mathbf{P}^*= \text{diag}(\mathbf{r}_u) \hspace{0.5em} e^{-\lambda \mathbf{M} \mathbf{Z}^\top} \text{diag}(\mathbf{r}_v),
\end{equation}
 where $e^{(\cdot)}$ is the element-wise exponential, and $\mathbf{r}_u$ and $\mathbf{r}_v$ are renormalization vectors in $\mathbb{R}^K$ and $\mathbb{R}^B$, respectively, which are calculated efficiently through a small number of matrix multiplications using the iterative Sinkhorn-Knopp algorithm~\cite{sinkhorn1967concerning}. We refer to \cite{cuturi2013sinkhorn} for derivation details. 

\paragraph{Joint optimizing online Sinkhorn-Knopp assignment and diffusion.} In our online bootstrapping method, we generally combine the diffusion training loss and the Sinkhorn-Knopp loss as follows:
\begin{equation}
\mathcal{L}_\text{online} = \mathcal{L}_{d} + \mathcal{L}_\text{SK}.
\end{equation}


During the initial steps of training, challenges arise concerning feature representativeness extracted from $v$, resulting in a potential conflict between $\mathcal{L}_{d}$ and $\mathcal{L}_\text{SK}$ losses. To address this, we implement several key modifications. Firstly, we introduce a warm-up coefficient, $\sigma$, to control the influence of the Sinkhorn-Knopp loss, $\mathcal{L}_\text{SK}$, during the warmup stage that emphasizes more unconditional training for a good initialization of diffusion features, the second stage, we focus on training the classifier-free guidance generation, utilizing signals from the prototype $\mathbf{M}$. Secondly, we use 
\texttt{stop-gradient} operation after extracting the features from diffusion models.  Lastly, we incorporate the moving average of the U-ViT model to extract the feature.


Based on previous analysis,  we design our diffusion loss $\mathcal{L}_{d}$ as follows:  
\begin{equation}
\label{eq:fm-obj-constant}
\mathcal{L}_{d} = \bbE_{t, \bx_t} ||\bx_1 - \bx_0 - v_\theta(\bx_t, t,{\bc} ) ||^2_2 ,
\end{equation}
where in the warmup stage, we set ${\bc} = \mathbf{0}$ as the zero vector to enable unconditional generation. In the main training stage, $\bc$ is extracted from the moving average of the diffusion models itself to achieve classifier-free guidance training. The condition signal $\bc$ is set as $\bc = \mathbf{m}^* $, or with a chance of $p_\text{drop} = 15\%$ to be dropped out to a learnable null embedding $\emptyset$, that helps to retain the ability of unconditioned generation.
\begin{equation}
\label{eq:get_condition}
    \mathbf{m}^* = \argmax_{\mathbf{m} \in \mathbf{M}}  \hspace{0.3em} \mathbf{m}^\top h_\phi\left({v}_{\bar{\theta}}^{l}(x_{t_{s}}, t_{s},\mathbf{0})\right).
\end{equation}
where $\bar{\theta}$ stands for the exponential moving average of the network parameter. $l$ and $t_s$ are hyperparameters used to extract features from diffusion models. $h_\phi({v}_{\bar{\theta}}^{l}(x_{t_{s}}, t_{s},\mathbf{0}))$ is  the image feature embedding $\bz$ as discussed above.

Our training process, detailed in Algorithm~\ref{alg:main}, incorporates these changes.  During optimization, several parameter sets are fine-tuned, including $\theta$ for the diffusion model, $\phi$ for the fully connected (FC) adaptor layer, $\bM$ for the learnable Sinkhorn-Knopp prototypes, and $\emptyset$ for the null embedding used in the classifier-free guidance.

\begin{algorithm}
\small
\caption{
Bootstrapping diffusion features for guidance by optimal transport assignment.
}
\label{alg:main}
\begin{algorithmic}[1]
\STATE{\textbf{Input}: $\bx_1$ the real data;  $v$ and $\theta$ the vector field predictor with parameters;  $h$ and  $\phi$ the FC adapter and parameters; $\mathbf{M}$ the learnable prototypes; and $\emptyset$ the learnable null embedding for classifier-free guidance.}
\STATE{\textbf{Parameters}: $N$ the number of iterations; $\sigma$ the threshold for warmup, default 0.5; $t_s$ the timestep and  $l$ the layer index for online self-guidance sampling; and $p_\text{drop}$ the dropout rate.}

\FOR{$iter=1,2,...,N-1$}
\STATE{Sample $\bx_0 \sim \mathcal{N}(0,1)$ from the Gaussian distribution, $\Tilde{\bx}_0 = \bx_0$ at $\hat{t}=0$.}
\STATE{Sample $\bx_1$.}
\IF{$iter/N < \sigma$}
\STATE{$\mathcal{L}_{d}  \leftarrow  \bbE_{t, \bx_t} ||\bx_1 - \bx_0 - v_\theta(\bx_t, t,\mathbf{0}) ||^2_2 $}
\ELSE
\STATE{Sample $\Tilde{p} \sim \texttt{Bernoulli}(p_\text{drop})$.}
\IF{$\Tilde{p}=0$}
\STATE{Obtain $\mathbf{m}^*$ from  Eq.~\ref{eq:get_condition}.}
\STATE{$\mathcal{L}_{d} \leftarrow \bbE_{t, \bx_t} ||\bx_1 - \bx_0 - v_\theta(\bx_t, t,\mathbf{m}^*) ||^2_2 $ }
\ELSE
\STATE{$\mathcal{L}_{d} \leftarrow \bbE_{t, \bx_t} ||\bx_1 - \bx_0 - v_\theta(\bx_t, t,\emptyset) ||^2_2 $.}
\ENDIF
\ENDIF
\STATE{Calculate $\mathcal{L}_\text{SK}$ from Eq.~\ref{eq:assign}.}
\STATE{$\mathcal{L} \leftarrow  \mathcal{L}_d + \min(\frac{iter}{\sigma \times N}, 1.0) \times \mathcal{L}_\text{SK}$.}
\STATE{Update $\theta,  \phi, \mathbf{M}, \emptyset$}
\ENDFOR
\STATE{\textbf{Return}:  $\theta,  \phi, \mathbf{M}, \emptyset$.}
\end{algorithmic}
\end{algorithm}

\paragraph{Sampling.}

We use a rounding procedure~\cite{asano2019selflabelling,caron2020unsupervised_swap} to obtain a discrete code, which is then applied in the sampling process for bootstrapping guidance. The guidance is governed by the equation:
\begin{equation}
{
v_\theta^\text{sample}(\bx_t, t, \bc) = v_\theta(\bx_t, t, \bc) +
g \cdot \left(v_\theta(\bx_t, t, \bc) - v_\theta(\bx_t, t, \emptyset)\right),
}
\label{eq:cfg}
\end{equation}
where $g$ represents the guidance strength that balances diversity and fidelity, and $\bc$ here is a prototype from $\bM$.

Sampling follows the ODE in Equation~\ref{eq:ode-lagrangian}. Utilizing estimations from Equation~\ref{eq:fm-obj} and Equation~\ref{eq:cfg}, we perform backward sampling by setting $\hat{\bx}_0 = \int_1^0 v_\theta^\text{sample}(\bx_t, t,\bc) \mathrm{d}t$, given that we have access to the noise $\bx_1 \sim p_1$. We solve this integration using numerical integrators.

For our guidance signal $\mathbf{c}$, we offer the flexibility to predefine its semantic meaning through visualization, as demonstrated in Appendix Section~\ref{supp:more_vis}, where different prototypes embody distinct meanings. Alternatively, we can condition $\mathbf{c}$ on a given image, leveraging a single forward pass of the network to query the most similar prototype from $\mathbf{M}$, and subsequently continue to condition our model on the queried prototype. This inherent flexibility is a defining feature of our framework, enhancing its adaptability and utility.

\vspace{-5pt}
\section{Experiments}

\subsection{Dataset, training and evaluation}

\paragraph{Datasets.} We report results on three datasets:  ImageNet256~\cite{imagenet}, ImageNet256-100~\cite{tian2020contrastive} (a subset of ImageNet-1k with 100 classes), and LSUN-Churches~\cite{yu15lsun}. The ablation study  is conducted on ImageNet100.
We train all methods for a total of 40,000, 100,000, and 300,000 iterations on ImageNet256-100, LSUN-Churches, and ImageNet256, respectively. This training takes approximately 2 days for ImageNet256-100, 4 days for LSUN-Churches, and 12 days for ImageNet256, utilizing 4 A1000 GPUs. All methods are compared under the same hyperparameter settings (see Appendix~\Cref{supp:hyperparameter}).

\paragraph{Training.}
We utilize the U-ViT-Large model from U-ViT~\cite{uvit}, which features 21 layers, 1 024 embedding dimensions, 4,096 MLP size, 16 heads, and 287M parameters. When provided with an image denoted as $I \in \mathbb{R}^{256\times256\times3}$, we conduct operations in the latent-space $z \in \mathbb{R}^{32\times32\times4}$.  For sampling the prototype $\bM$, we closely adhere to the distribution observed in the data. This ensures that our sampling methodology is representative and grounded in the actual data characteristics.  
We use $\sigma{=}0.5$ during training, and the Exponential Moving Average (EMA) is open till the Sinkhorn-Knopp loss finishes its warmup stage. When calculating the Sinkhorn-Knopp loss during the warmup stage, we mask the timestep $t$ within the range [0.15, 0.25] to ensure that the network focuses on this specific time step for improved discriminative performance.
We use the \texttt{dopri5} ODE solver for sampling, and guidance strength $g{=}0.4$.
For more detailed implementation specifics we refer to the Appendix. Code will be released.

\paragraph{Evaluation.}
We evaluate both diversity and fidelity of the generated images by the Fr\'echet Inception Distance (FID)~\citep{heusel2017gans_fid}, as it is the most common metric for the evaluation of generative methods, e.g.,~\citep{dhariwal2021diffusion_beat,karras2019_stylegan,sgdm,saharia2022photorealistic_imagen}. It provides a symmetric measure of the distance between two distributions in the feature space of Inception-V3~\citep{szegedy2016rethinking_inceptionv3}. 
We compute the FID using 10,000 reference images and use it as our main metric for the sampling quality. We also measure the Inception Score (IS)~\citep{salimans2016improved_is_inceptionscore}, following common practice~\citep{dhariwal2021diffusion_beat,karras2019_stylegan,brock2018large_biggan,sgdm}. IS measures how well a model fits into the full ImageNet class distribution.

\subsection{Results}

Our experiments comprise three main parts. Firstly, we investigate the impact of offline guidance, examining its effects concerning the timestep $t$ and the layer index of U-ViT. Secondly, we present results from our online self-guidance experiments on ImageNet and LSUN-Churches, accompanied by several ablation studies. Lastly, we offer a range of analyses and visualizations to enhance our method's comprehension.


\begin{figure}
\centering
\begin{subfigure}{0.5\textwidth}
  \centering
  \includegraphics[width=.97\textwidth]{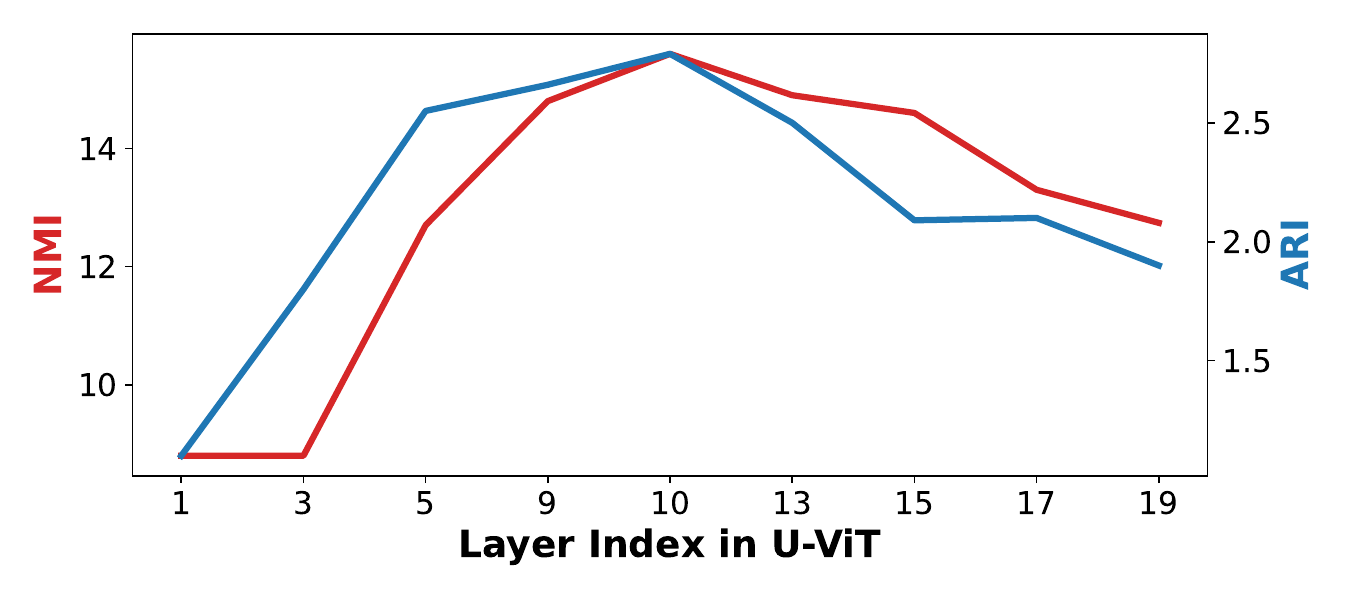}
  \label{fig:sub1}
\end{subfigure}%
\newline
\begin{subfigure}{.5\textwidth}
  \centering
  \includegraphics[width=.97\textwidth]{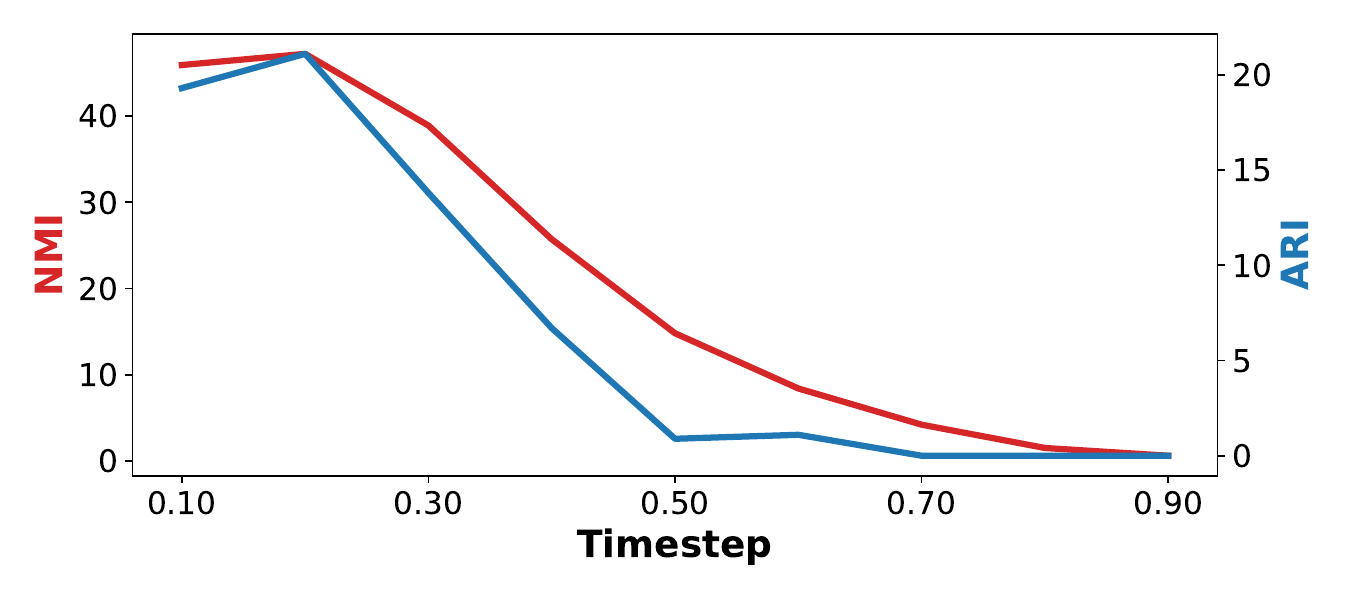}
  \label{fig:sub2}
\end{subfigure}
 \vspace{-23pt}
\caption{\textbf{Performance in clustering across various layer numbers and timesteps.} Timestep 0.2 and layer index 10 is the optimal combination. }
\label{fig:ablation}
\end{figure}

\vspace{-10pt}
\paragraph{Ablation studies about offline guidance.}
\label{sec:ablation_offline}


We conduct ablation studies on offline guidance signal extraction, considering two key factors: the timestep $t$ and the layer index of U-ViT.
Figure~\ref{fig:ablation} showcases the results of our ablation study. Following~\cite{sgdm}, we employ Normalized Mutual Information (NMI) and Adjusted Rand Index (ARI) metrics to assess the effectiveness of our guidance signal in terms of clustering, with our ultimate aim being to improve fidelity. The study indicates that the optimal performance in terms of NMI and ARI is achieved at the midpoint of the model's layers (specifically, the $10^\text{th}$ layer in a 21-layer structure) and at the timestep of 0.2. 
%

\paragraph{Ablation studies about bootstrapping guidance.}
\label{sec:ablation_online}
We draw the conclusion of the optimal layer index and timestep with offline guidance, and apply this setup to our online guidance experiments, using layer index 10 and timestep 0.2. 
In addition, we conduct several ablation studies on the online bootstrapping learning paradigm, exploring key factors that include the necessity of \texttt{stop-gradient} to mitigate interaction between the diffusion loss and the Sinkhorn-Knopp loss, the impact of the warm-up weight (which determines reliance on a well-developed diffusion representation before applying the Sinkhorn-Knopp loss), the use of the \texttt{[CLS]} token instead of using average-pooled embeddings from patches, and the importance of masking timesteps within the range [0.15, 0.25] during bootstrapping guidance extraction to focus the network on this specific timeframe. Our findings, summarized in Table~\ref{tab:ablation}, emphasize the significance of these choices in optimizing FID on ImageNet256-100.
%



\begin{table*}[h]
    \begin{minipage}{.6\linewidth}
    \centering
        \resizebox{\textwidth}{!}{
    \begin{tabular}{c>{\columncolor{mygray}}c>{\columncolor{mygray}}ccc}
    \toprule
      & \textit{anno-free} & \textit{online} & FID $\downarrow$  & IS $\uparrow$   \\
      \hline
       Unconditional & \textbf{--} & \textbf{--} & 45.1& 26.5 \\
        \hline
      Offline Guidance from diffusion $^\dag$ & \cmark & \xmark &35.8 & {34.8}\\
        Bootstrapping guidance &  \cmark & \cmark&36.0& 34.3 \\
        Bootstrapping guidance (Patch) & \cmark &\cmark &33.4& 45.3 \\
       \hline
       Class guidance  $^\ddag$ & \xmark & \textbf{--} &34.4& 44.1\\
      DINO guidance $^\ast$ &  \cmark & \xmark & {32.0}& {48.7}\\
       \bottomrule
    \end{tabular}
    }
     \vspace{-10pt}
    \caption{\textbf{Main result on ImageNet256-100}. $^\dag$ requires pretrained diffusion model. $^\ddag$ requires GT labels as heuristics. $^\ast$ requires off-the-shelf pretrained DINO encoder.}
    \label{tab:main_in100}    
    \end{minipage}
    \hspace{1em}
    \begin{minipage}{.37\linewidth}
    \centering
        \resizebox{.9\textwidth}{!}{
    \begin{tabular}{lcc}
    \toprule
        & FID $\downarrow$ & IS $\uparrow$ \\
      \hline
      \textbf{Our Method} &36.0&34.3\\
      \hline
      w/o stop-gradient &{45.0} &{24.1} \\
     w/o \texttt{[CLS]} token &{37.1} &{33.1}\\
     w/o warmup weight &{46.1} &{22.1} \\
     w/o masking timestep &{39.8} &{31.2} \\
       \bottomrule
    \end{tabular}
    }
    \vspace{-7pt}
    \caption{\textbf{Ablation study about bootstrapping guidance on ImageNet256-100.} All four choices prove to be advantageous. 
    }
    \label{tab:ablation}
    \end{minipage} 
\end{table*}



\begin{table}
    \centering
    \resizebox{.5\textwidth}{!}{
        \begin{tabular}{cc@{\hspace{.7em}}ccc}
            \toprule
            & \multicolumn{2}{c}{\textit{ImageNet256}} & \multicolumn{2}{c}{\textit{LSUN-Churches}} \\
            \cmidrule(lr){2-3} \cmidrule(lr){4-5} 
            & FID $\downarrow$  & IS $\uparrow$   & FID $\downarrow$  & IS $\uparrow$ \\
            \hline
            Unconditional  & 42.5 & {23.1} & 13.10 & {2.95} \\
            \hline
            Offline guidance from diffusion $^\dag$ & 29.8 & 38.1 & 12.75 & {3.03} \\
            Bootstrapping guidance & 30.9 & 32.3 & 12.38 & 3.17 \\
            \hline
            Class Guidance $^\ddag$  & 19.1 & {61.2} & -- &  -- \\
            \bottomrule
        \end{tabular}
    }
    \vspace{-10pt}
    \caption{\textbf{Main results on ImageNet256 and LSUN-Churches}. $^\dag$ with pretrained diffusion model. $^\ddag$ requires ground truth labels.}
    \label{tab:main_in256_churches}
\end{table}

\begin{figure*}
    \centering
\includegraphics[width=0.9\textwidth]{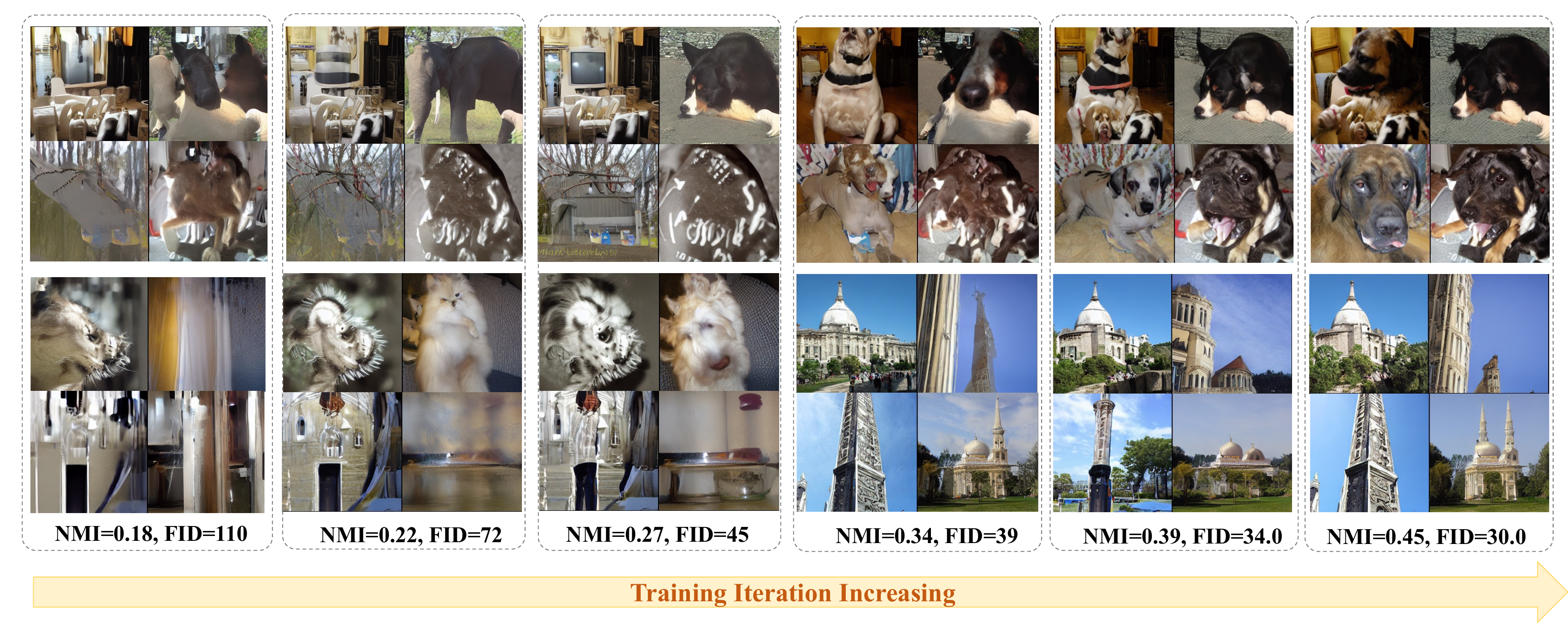}
\vspace{-12pt}
    \caption{\textbf{The trend of Normalized Mutual Information (NMI) and FID during training.} Our generated images initially exhibit low discriminability (low NMI) and suboptimal fidelity in the first half of iterations but gradually improve in fidelity (FID) and cluster cohesiveness (high NMI) during the latter half of iterations.}
\label{fig:clustervis_during_training}
\end{figure*}

\vspace{-10pt}
\paragraph{Main results.}


Incorporating our prior research findings in offline and online guidance, we present our experimental results across three distinct datasets: ImageNet256-100 (Table~\ref{tab:main_in100}), ImageNet256, and LSUN-Churches (Table~\ref{tab:main_in256_churches}).
We primarily compare our approach with unconditional diffusion models, class-conditioned diffusion models, and DINO guidance. For offline guidance, we employ 5,000, 300, and 30 feature clusters for ImageNet256, ImageNet256-100, and LSUN-Churches, respectively. For DINO guidance, we utilize features from DINO~\cite{dino} by applying 300 clusters using $k$-means  the one-hot embeddings are applied as the offline guidance, following a similar approach to~\cite{sgdm}. In the case of class-conditional diffusion models, we rely on existing labels. To ensure a fair comparison, we maintain consistent training hyperparameters and model sizes across all experiments.


For ImageNet result in Table~\ref{tab:main_in100} and ~\ref{tab:main_in256_churches} (left), our proposed approaches of offline self-guidance and online self-guidance both significantly outperform unconditional diffusion models. 
DINO~\cite{dino} slightly outperforms our online bootstrap method, suggesting potential improvements in the interplay between guidance signals and diffusion models for bootstrapping. However, it's important to acknowledge that a performance gap persists when compared to class-conditioned diffusion models, which heavily depend on extensive data annotations.

We also explore the extension of patch-level bootstrapping guidance, inspired by recent advancements in self-supervised learning~\cite{caron2022location,ziegler2022self}. The only major difference from our aforementioned image-level bootstrapping guidance lies in using individual patch tokens instead of the \texttt{[CLS]} token in U-ViT, followed by the patch-level Sinkhorn-Knopp assignment rather than on images. In~\Cref{tab:main_in100}, we observe that the patch-level guidance leads to better fidelity than image-level bootstrapping guidance and class guidance, while the gap with DINO guidance is marginal. This more detailed guidance at the level of patches can yield better fidelity, which is in line with the observations from the literature~\cite{sgdm,zhou2111lafite}.

In Table~\ref{tab:main_in256_churches} (right), our results on LSUN-Churches indicate that both our offline and online self-guidance approaches outperform unconditional generation, with the online self-guidance yielding the best performance. 

\vspace{-10pt}
\paragraph{Training dynamics.}
In Figure~\ref{fig:clustervis_during_training}, we showcase the progression of NMI and FID during training. As training advances, both NMI and FID show consistent improvement, indicating that both feature discriminativeness and generation quality are enhanced over time, underscoring the effectiveness of our online bootstrapping method. Notably, during the warmup stage (the left half of the figure), the network predominantly focuses on unconditional generation, resulting in random images. 
In contrast, the model gradually adapts to generate images that align within a cluster in the later stage of training. Additional results of losses, NMI, and FID curves during training are available in Appendix~\Cref{supp:more_exps}.

\paragraph{Visualization of image-level and patch-level guidance.} 
Given a query image to condition our model on, we can generate multiple images that belong to the same cluster as the query, as illustrated in Figure~\ref{fig:cluster_vis_by_query}. We further show the clustering of the online self-guidance in Appendix Section~\ref{supp:more_exps}. 
In Figure~\ref{fig:patchlevel_cluster_vis}, we visualize sampled images using patch-level bootstrapping guidance. It demonstrates that the generated images align well in layout with individual patches of the query images, highlighting the flexibility of our method to transition from image-level to patch-level guidance.

\begin{figure}
    \centering
    \includegraphics[width=0.5\textwidth]{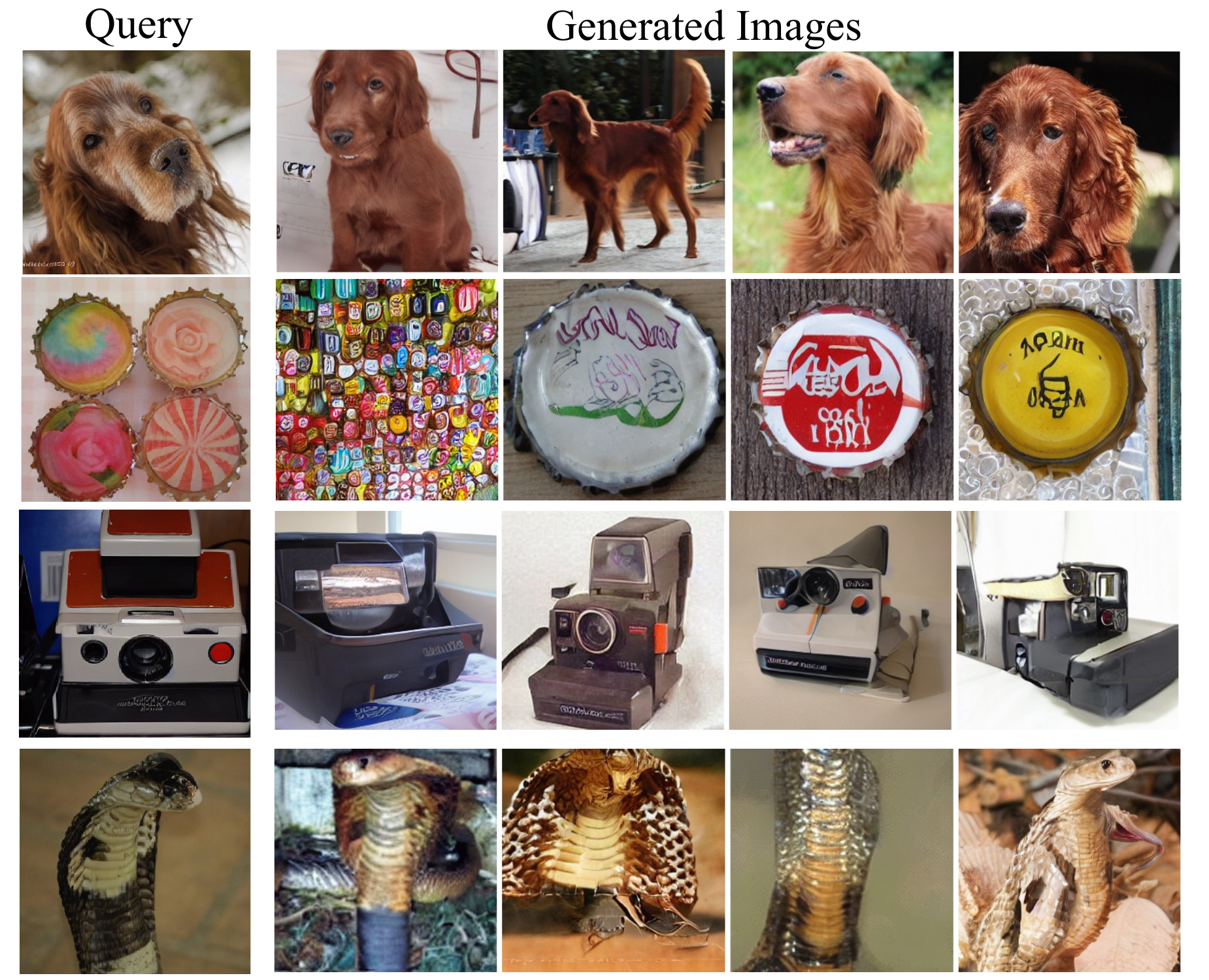}
    \vspace{-20pt}
    \caption{Using a query image in the left column for \textbf{image-level guidance}, we generate multiple images with the same prototype features.  The generated images share consistent semantic meaning while exhibiting diverse appearances.}
    \label{fig:cluster_vis_by_query}
\end{figure}

\begin{figure}
    \centering
    \includegraphics[width=0.5\textwidth]
    {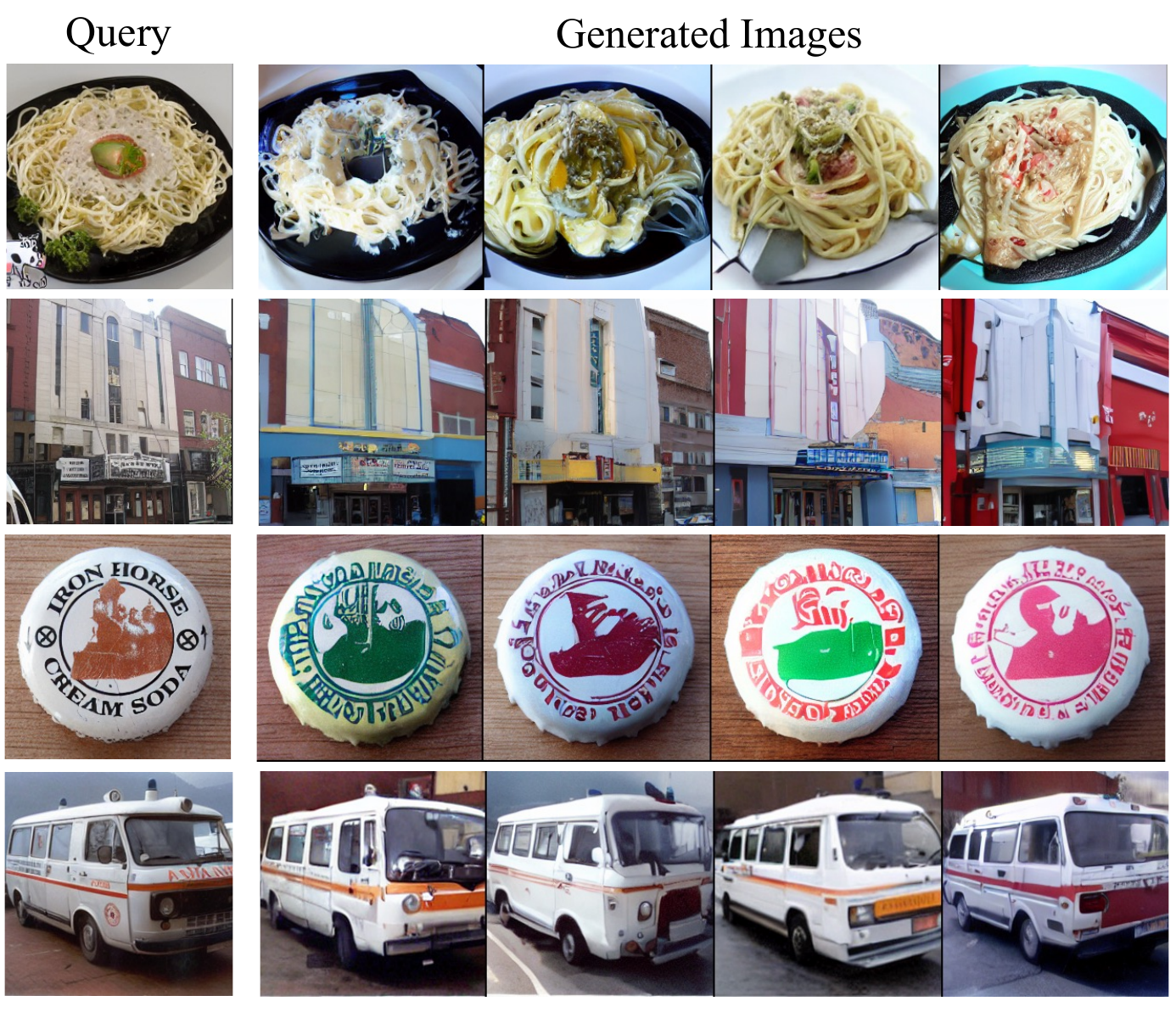}
    \vspace{-20pt}
    \caption{Using a query image in the left column for \textbf{patch-level guidance}, we generate multiple images with the same prototype features. The generated images maintain a consistent semantic layout while featuring diverse backgrounds and slight variations.}
    \label{fig:patchlevel_cluster_vis}
\end{figure}

\begin{figure}
\begin{subfigure}{.49\textwidth}
  \centering
  \includegraphics[width=.98\linewidth]{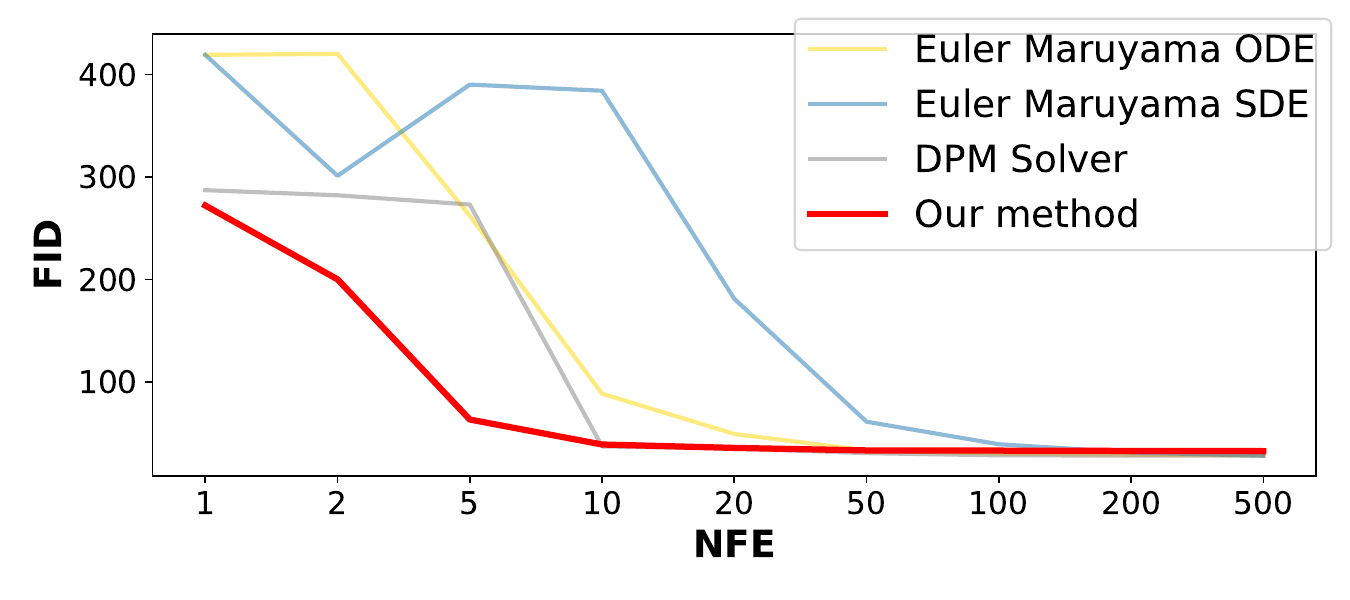}
  \vspace{-10pt}
  \caption{\textbf{FID \textit{v.s.} NFE on ImageNet100.}}
  \label{fig:fid_vs_nfe}
\end{subfigure}%
\newline
\begin{subfigure}{.49\textwidth}
  \centering
  \includegraphics[width=.98\linewidth]{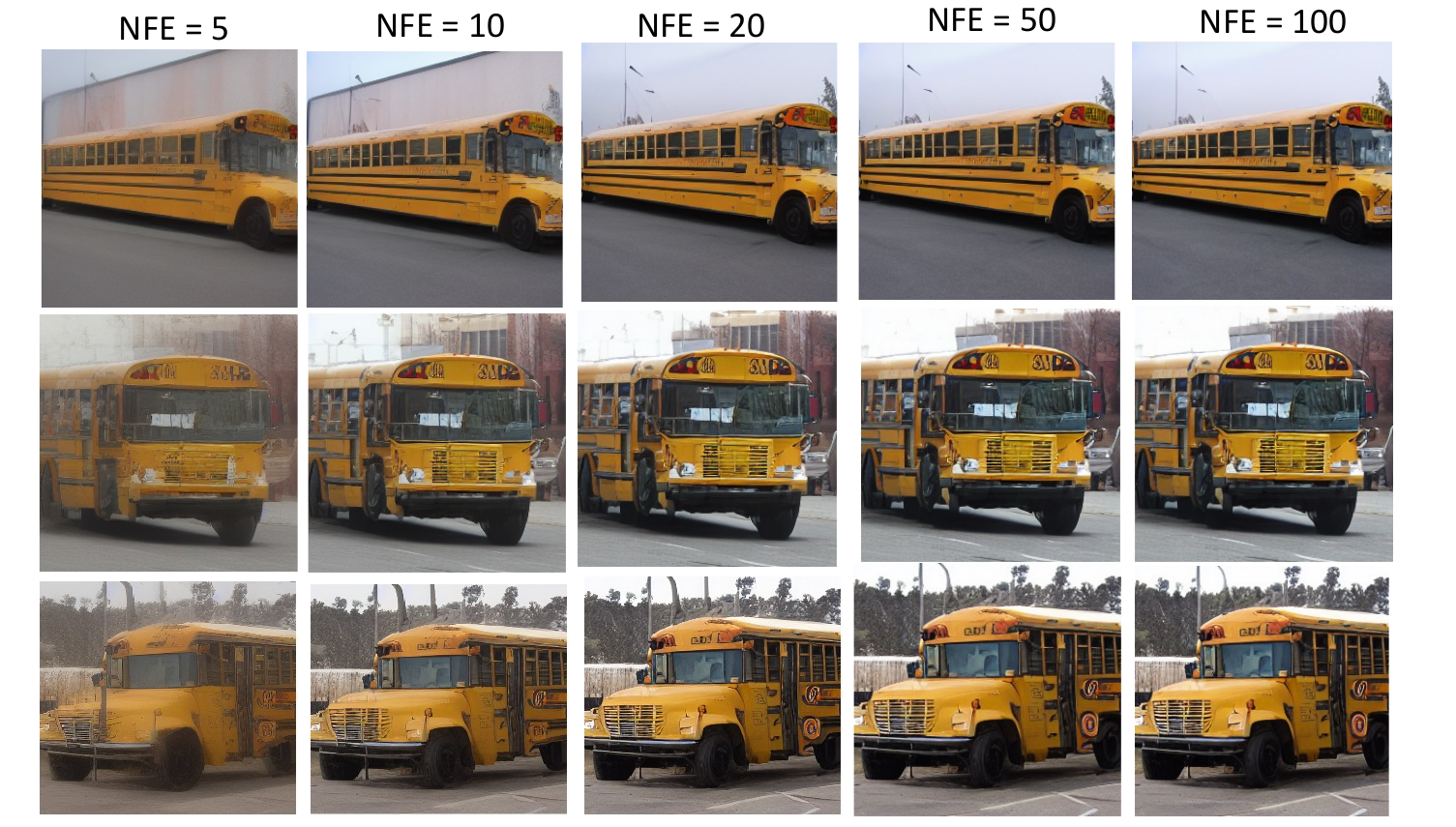}
    \caption{\textbf{Visualizing clusters at different sampling steps.}}
    \label{fig:vis_sampling_steps}
\end{subfigure}
\vspace{-14pt}
\caption{\textbf{Results of different sampling steps.} Our method can achieve better sampling efficiency.}
\end{figure}

\vspace{-10pt}
\paragraph{Samping steps.}
To dissect the sampling efficiency of our method, we  primarily compare with three sampling baselines: Euler-Maruyama ODE~\citep{song2021scorebased_sde} and Euler-Maruyama SDE~\citep{song2021scorebased_sde}, and DPM solver~\citep{lu2022dpmsolver}. We focus on evaluating the trade-off between Fréchet Inception Distance (FID) and the Number of Function Evaluations (NFE), as illustrated in Figure~\ref{fig:fid_vs_nfe}. Notably, our method demonstrates a superior trade-off between sampling steps and fidelity, particularly in the low-NFE regime. Additionally, we provide visualizations of different sampling steps in Figure~\ref{fig:vis_sampling_steps}. It is evident that sampling in just 20 steps already yields vivid results, and further increasing the number of sampling steps leads to diminishing improvements in sampling quality. This demonstrates our framework's sampling efficiency.

\section{Conclusion}


In this paper, we introduce two frameworks for exploring self-guidance within diffusion models, allowing us to leverage their features either offline or through online self-bootstrapping. Our methods unlock the latent potential of diffusion models during training, surpassing unconditional generation and performing comparably to class-conditioned generation. In the future,  we would like to explore other similar backbone such as DiT~\cite{dit_peebles2022scalable} to validate the efficacy of our method.


\clearpage
{
    \small
    \bibliographystyle{ieeenat_fullname}
    \bibliography{main}
}

\clearpage

\maketitlesupplementary
\onecolumn

\tableofcontents

\section{Hyperparameters}
\label{supp:hyperparameter}

\paragraph{Training Hyperparameters}  In practice, we observe that using only $3$ iterations in Sinkhorn-Knopp Algorithm is fast and sufficient to obtain good performance. 

In offline guidance, as the pretrained weight from U-ViT is trained with classifier-free guidance, we set the conditional embedding to null to facilitate feature extraction.

Please refer to Table~\ref{supp:training_details} for additional training hyperparameters, and to Table~\ref{supp:training_device_and_time} for details about the training device and time. 


\begin{table}
\begin{center}
\scalebox{0.88}{
\begin{tabular}{lccc}
\toprule
    Dataset & ImageNet100$\times$256 & ImageNet$\times$256 & LSUN-Churches$\times$256 \\
    \midrule
    Latent space  & $\checkmark$ & $\checkmark$ & $\checkmark$ \\
    Latent shape &  32$\times$32$\times$4 & 32$\times$32$\times$4 & 32$\times$32$\times$4 \\
    Image decoder &  ft-EMA & ft-EMA & ft-EMA \\
    \midrule
    Batch size & 512 & 512 & 512 \\
    Training iterations & 40K & 300K & 100k \\
    Warm-up steps & 5K & 5K & 5K \\
    \midrule
     Cluster Number & 300 & 5{,}000 & 30 \\
    \midrule
    Optimizer & AdamW & AdamW & AdamW \\
    Learning rate &  5e-5 & 5e-5 & 5e-5 \\
    Weight decay & 0.00 & 0.00 & 0.00 \\
    Betas  & (0.9, 0.999) & (0.9, 0.999) & (0.9, 0.999) \\
    \midrule
    Noise schedule & OT & OT & OT \\
    \midrule
    Sampler  & dopri & dopri & dopri \\
    \midrule
    CFG & $\checkmark$ & $\checkmark$ & $\checkmark$ \\
    $p_{\mathrm{uncond}}$ &  0.1 & 0.1 & 0.1 \\
    Guidance strength & 0.4 & 0.4 & 0.4 \\
    \midrule
    Convolution & $\times$ & $\times$ & $\times$\\
    \bottomrule
\end{tabular}}\vspace{-.2cm}
\end{center}
\caption{The experimental setup for U-ViT in the main paper. ``ft-EMA" and ``original" correspond to different weights of the image decoder provided in \url{https://huggingface.co/stabilityai/sd-vae-ft-ema}.  ``OT'' denotes the OT path in flow matching~\cite{lipman2022flow,rectifiedflow_iclr23}.``$p_{\mathrm{uncond}}$" represents the unconditional training probability in classifier free guidance (CFG). ``Convolution" represents whether to add a 3$\times$3 convolutional block before output.}
\label{supp:training_details}
\end{table}

\begin{table}
    \centering
    \scalebox{0.95}{
    \begin{tabular}{ccccc}
    \toprule
    Dataset & Model & Training devices & Training time & Training iterations \\
    \midrule
    ImageNet100$\times$256 & U-ViT-L/2 & 4 GeForce RTX A5000 & 48 hours & 40K \\
    \midrule
    LSUN-Churches$\times$256  & U-ViT-L/2 & 4 GeForce RTX A5000  & 96 hours & 100K \\
    \midrule
    ImageNet$\times$256 & U-ViT-L/2 & 4 GeForce RTX A5000 & 290 hours & 300K \\
    \bottomrule
    \end{tabular}}
    \caption{The training devices and time.}
    \label{supp:training_device_and_time}
\end{table}

\begin{figure}
    \centering
    \includegraphics[width=0.7\linewidth]{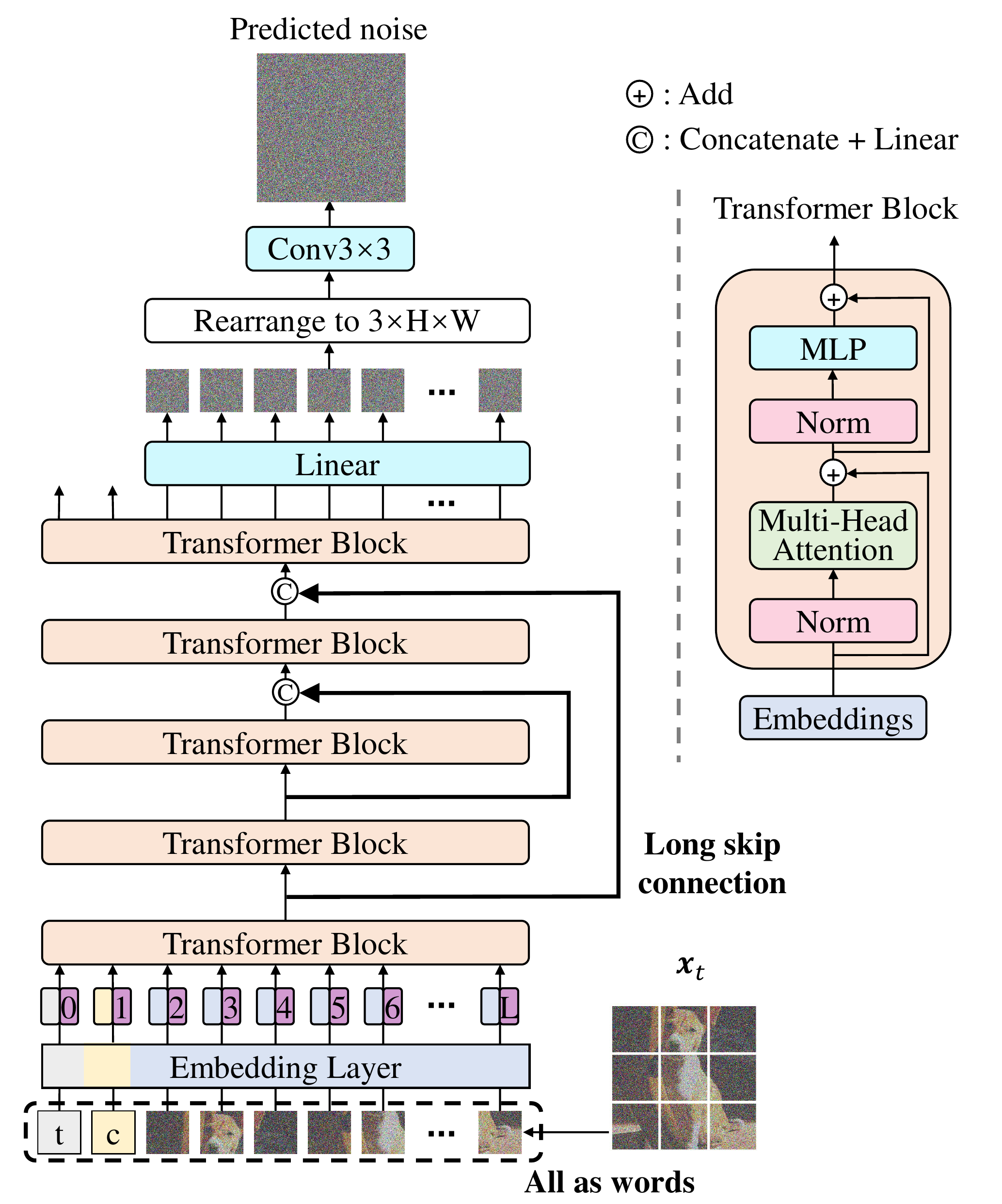}
    \caption{The \textbf{U-ViT} architecture for diffusion models, which is characterized by treating \textbf{all} inputs including the time, condition and noisy image patches \textbf{as tokens} and employing (\#Blocks-1)/2 \textbf{long skip connections} between shallow and deep layers. The network structure is not our contribution. For completeness, this figure is replicated from the paper~\cite{uvit}.}.\vspace{-.2cm}
    \label{fig:uvit}
\end{figure}

\paragraph{Network structures} We primarily investigate the network structure from U-ViT~\cite{uvit}. For completeness, we have replicated it in Figure~\ref{fig:uvit}.

\paragraph{Evaluation details} 
We use the common package Clean-FID~\citep{parmar2021cleanfid}, torch-fidelity~\citep{obukhov2020torchfidelity} for FID, IS calculation, respectively.
 For the checkpoint, we pick the checking point every 4,000 iterations by minimal FID between generated sample set and the train set.  We utilize the precomputed mean and standard deviation statistics from U-ViT~\cite{uvit}.





\paragraph{Other attempts} We also considered adding a queue of size 5{,}120, which is 10 times the batch size of 512. However, we have empirically found that it does not provide any additional benefits.

\section{Extra Related Works}
\label{supp:more_relatedworks}

\paragraph{SDE and ODE in Diffusion models.} 
Score-based generative model family features seminal models like Score Matching with Langevin Dynamics (SMLD) by Song et al.~\citep{song2019generative}, and DDPMs by Ho et al.~\citep{ho2020denoising}. Both SMLD and DDPMs are encompassed within the Stochastic Differential Equations (SDEs) framework, a concept further elaborated by Song et al.~\cite{song2021scorebased_sde}.
Recent advancements, as illustrated by Karras et al.~\cite{karras2022elucidating} and Lee et al.~\cite{lee2023minimizing}, have shown that utilizing an Ordinary Differential Equation (ODE) sampler for diffusion SDEs can significantly reduce sampling costs compared to traditional methods involving the discretization of diffusion SDEs. Additionally, within the context of Flow Matching regimes~\cite{lipman2022flow} and Rectified Flow~\cite{rectifiedflow_iclr23}, both SMLD and DDPMs can be considered as special cases under the Variance Preserving and Variance-Explosion paths of the Probability Flow ODE framework~\cite{song2021scorebased_sde}.
We focus on the ODE approach as opposed to the traditional SDE method, a pivot that promises new insights and efficiencies in the development of generative models. 

Several~\cite{lipman2022flow,rectifiedflow_iclr23} pointed out that using a non-linear interpolation for $\bx_t$ in the VP probability flow ODE in Equation~\ref{eq:vp-path} can result in an unnecessary increase in the curvature of generative trajectories.
As a consequence, this can lead to reduced efficiency in the training and sampling of generative trajectories.

\section{Method Details}
\label{supp:more_method}

\begin{figure}
\begin{subfigure}{.99\textwidth}
  \centering
  \includegraphics[width=.98\linewidth]{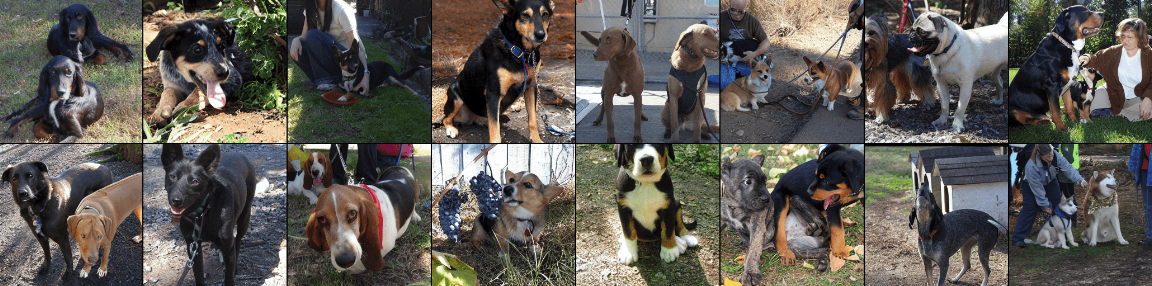}  
  \caption{Cluster \#3: ``dogs with dark color."}
  \label{fig:sub-second}
\end{subfigure}
\newline
\begin{subfigure}{.99\textwidth}
  \centering
  \includegraphics[width=.98\linewidth]{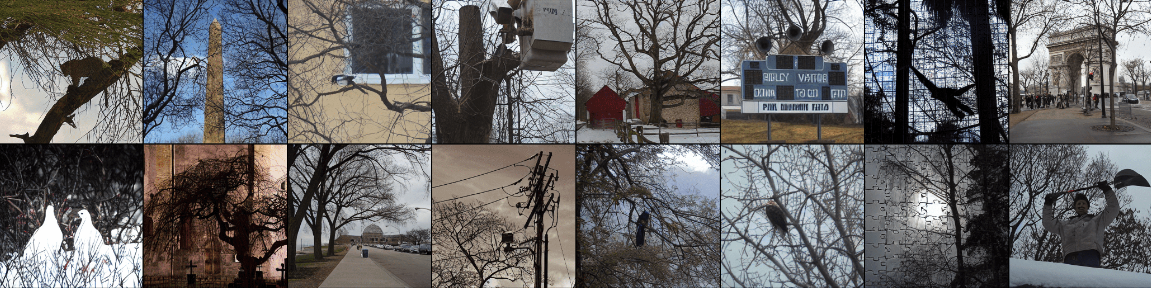}  
  \caption{Cluster \#8: ``A scene with branchy trees alongside a road.''}
  \label{fig:sub-second}
\end{subfigure}
\newline
\begin{subfigure}{.99\textwidth}
  \centering
  \includegraphics[width=.98\linewidth]{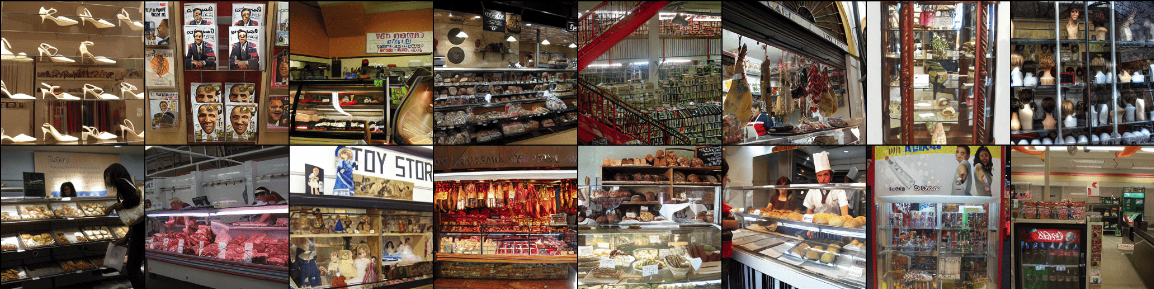}  
  \caption{Cluster \#10: ``A retail store displays a wide range of goods, from everyday necessities to treats and leisure items.''}
  \label{fig:sub-second}
\end{subfigure}
\caption{Semantic Interpretation: the visualization of ImageNet256 from various clusters.}
\label{fig:clustervis_in256}
\end{figure}

\begin{figure}
\begin{subfigure}{.99\textwidth}
  \centering
  \includegraphics[width=.98\linewidth]{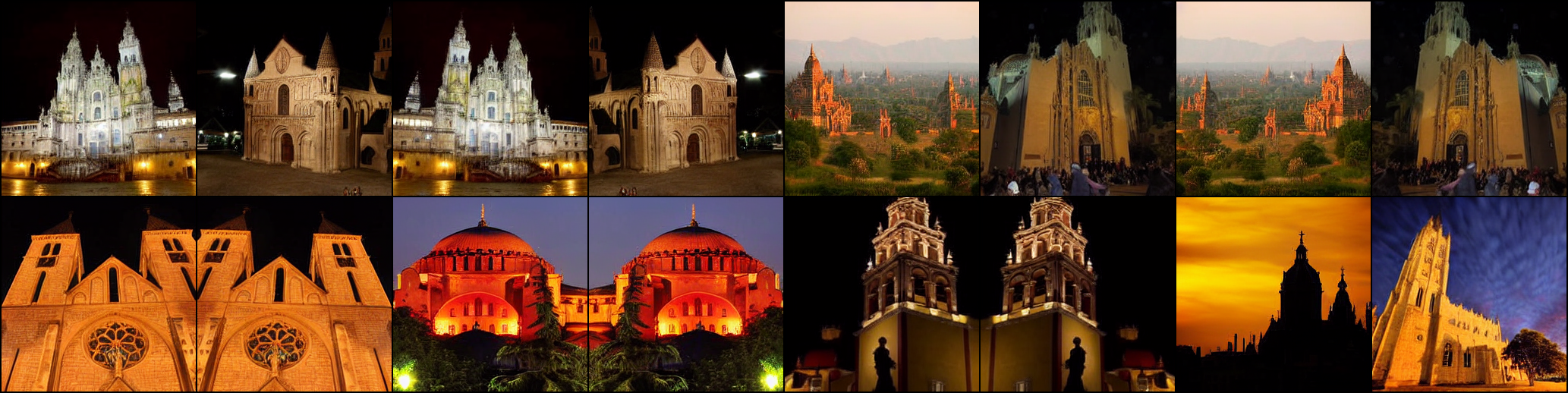}  
  \caption{Cluster \#1: ``cathedrals or large churches photographed at various times during the evening or night.''}
  \label{fig:sub-second}
\end{subfigure}
\newline
\begin{subfigure}{.99\textwidth}
  \centering
  \includegraphics[width=.98\linewidth]{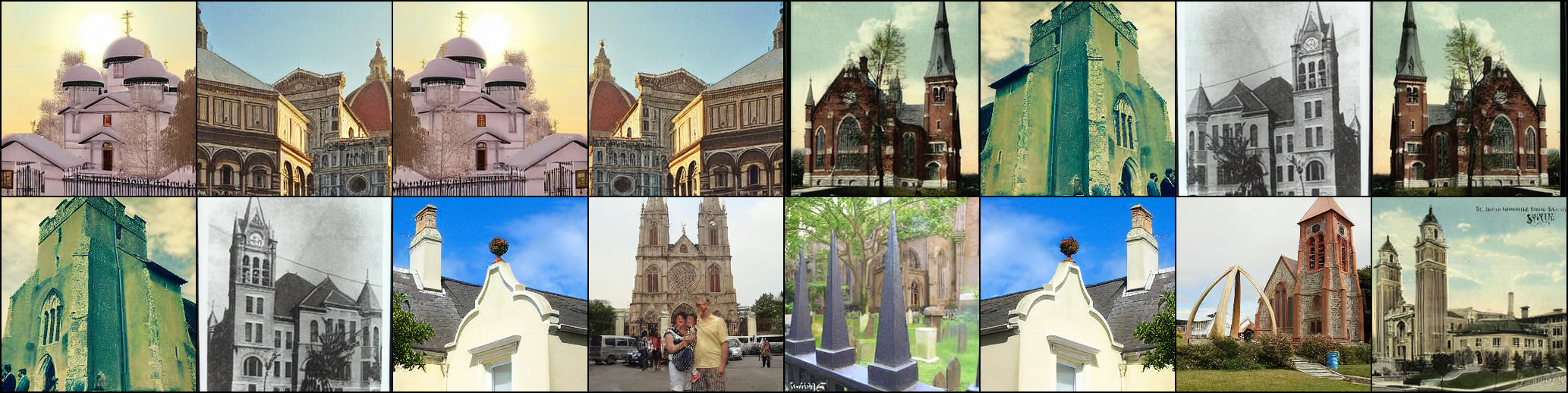}  
  \caption{Cluster \#4: ``Each image features church buildings, depicted using different artistic styles or filter effects. These effects provide a varied aesthetic to the images, suggesting an exploration of the churches through diverse visual interpretations, likely for artistic or illustrative purposes.''}
  \label{fig:sub-second}
\end{subfigure}
\newline
\begin{subfigure}{.99\textwidth}
  \centering
  \includegraphics[width=.98\linewidth]{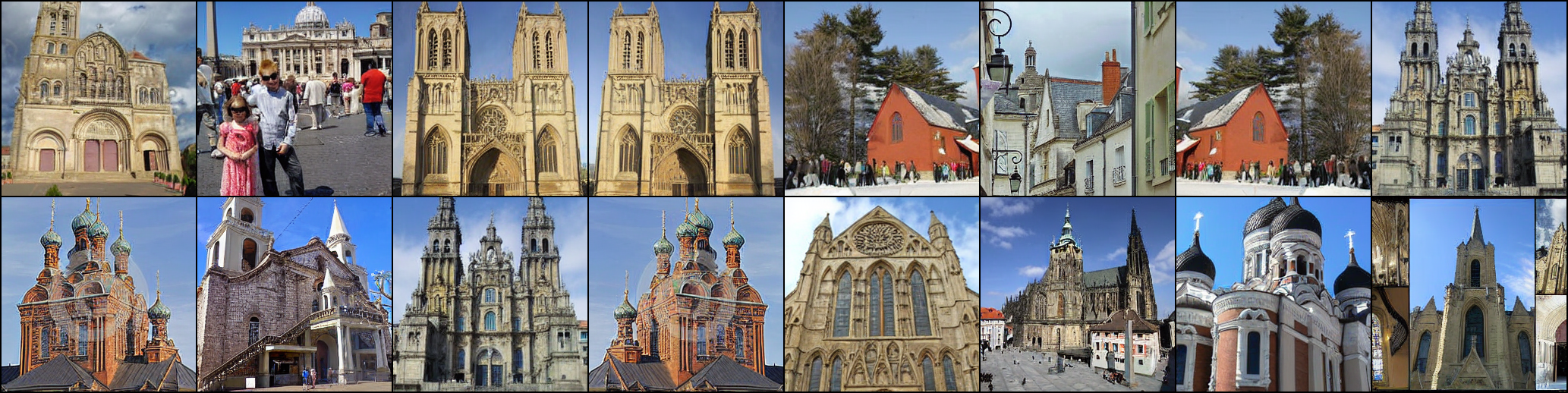}  
  \caption{Cluster \#7:``All feature prominent church buildings, which showcase a variety of architectural styles from different historical periods and cultural contexts.''}
  \label{fig:sub-second}
\end{subfigure}
\caption{Semantic Interpretation: the visualization of churches from various clusters out of 30.}
\label{fig:clustervis_churches}
\end{figure}

\paragraph{Semantic Assignment for self-learned clusters.} Sampling can be conducted under specific guidance signals. These samples can then be visualized and semantically interpreted, as demonstrated in Figure~\ref{fig:clustervis_in256} (ImageNet256) and Figure~\ref{fig:clustervis_churches} (LSUN-Churches).


\paragraph{ODE Sampling.} We illustrate the pseudo-code for ODE sampling in Algorithm~\ref{alg:ode_sampling_python}.

 \begin{algorithm}[H]
 \small
 \caption{
 Euler Sampling algorithm.
 }
 \label{alg:ode_sampling_python}
 \definecolor{codeblue}{rgb}{0.25,0.5,0.5}
 \definecolor{codekw}{rgb}{0.85, 0.18, 0.50}
 \lstset{
   backgroundcolor=\color{white},
   basicstyle=\fontsize{9.2pt}{9.2pt}\ttfamily\selectfont,
   columns=fullflexible,
   breaklines=true,
   captionpos=b,
   commentstyle=\fontsize{9.2pt}{9.2pt}\color{codeblue},
   keywordstyle=\fontsize{9.2pt}{9.2pt}\color{codekw},
   escapechar={|}, 
   xleftmargin=.02\textwidth, xrightmargin=.02\textwidth
 }
 \begin{lstlisting}[language=python]
 def ode_sampling(model, noise, x0):
   # x0: gaussin noise.
   # model: pretrained vector field predictor
   z = noise.detach().clone()
   dt = 1.0 / N
   est, traj = [], []

   for i in range(0, N, 1):  # fix-step Euler ODE solver
         t = i / N
         pred = model(z, t) # vector field prediction
         pred = pred.detach().clone()
         _est_now = z + (1 - t)*pred
         est.append(_est_now)

         z = z.detach().clone() + pred * dt
         traj.append(z.detach().clone())

     return traj[-1], est
 \end{lstlisting}
 \end{algorithm}

\paragraph{The training process of diffusion framework.} For the sake of completeness, we reiterate the algorithm of training outlined in~\Cref{alg:flow_matching_pseudo}.

\begin{algorithm}
        \caption{The training process of diffusion framework.}
 \label{alg:flow_matching_pseudo}
\begin{algorithmic}[1]
\STATE{\textbf{Input:} Empirical distribution $q_1$, gaussian distribution $q_0$,  batchsize $b$, initial network $v_{\theta}$.}

\WHILE{Training}
 \STATE{Sample batches of size $b$ \textit{i.i.d.} from the datasets}
 \STATE{$\vx_0 \sim q_0(\vx_0); \quad \vx_1 \sim q_1(\vx_1)$}
\STATE {$t \sim \mathcal{U}(0, 1)$}
\STATE{\textcolor{blue}{\texttt{\# Interpolation.}}}
\STATE {$x_t \gets t \vx_1 + (1 - t) \vx_0$}

\STATE {$\text{DM}(\theta) \gets \| v_\theta(x_t,t) - (\vx_1 - \vx_0)\|^2$}
\STATE {$\theta \gets \mathrm{Update}(\theta, \nabla_\theta \text{DM}(\theta))$}
 \ENDWHILE
\STATE {\textbf{Return} $\text{DM}$}

\end{algorithmic}
\end{algorithm}

\section{Experimental Results}
\label{supp:more_exps}

\begin{figure}
    \centering
    \includegraphics[width=.67\textwidth]{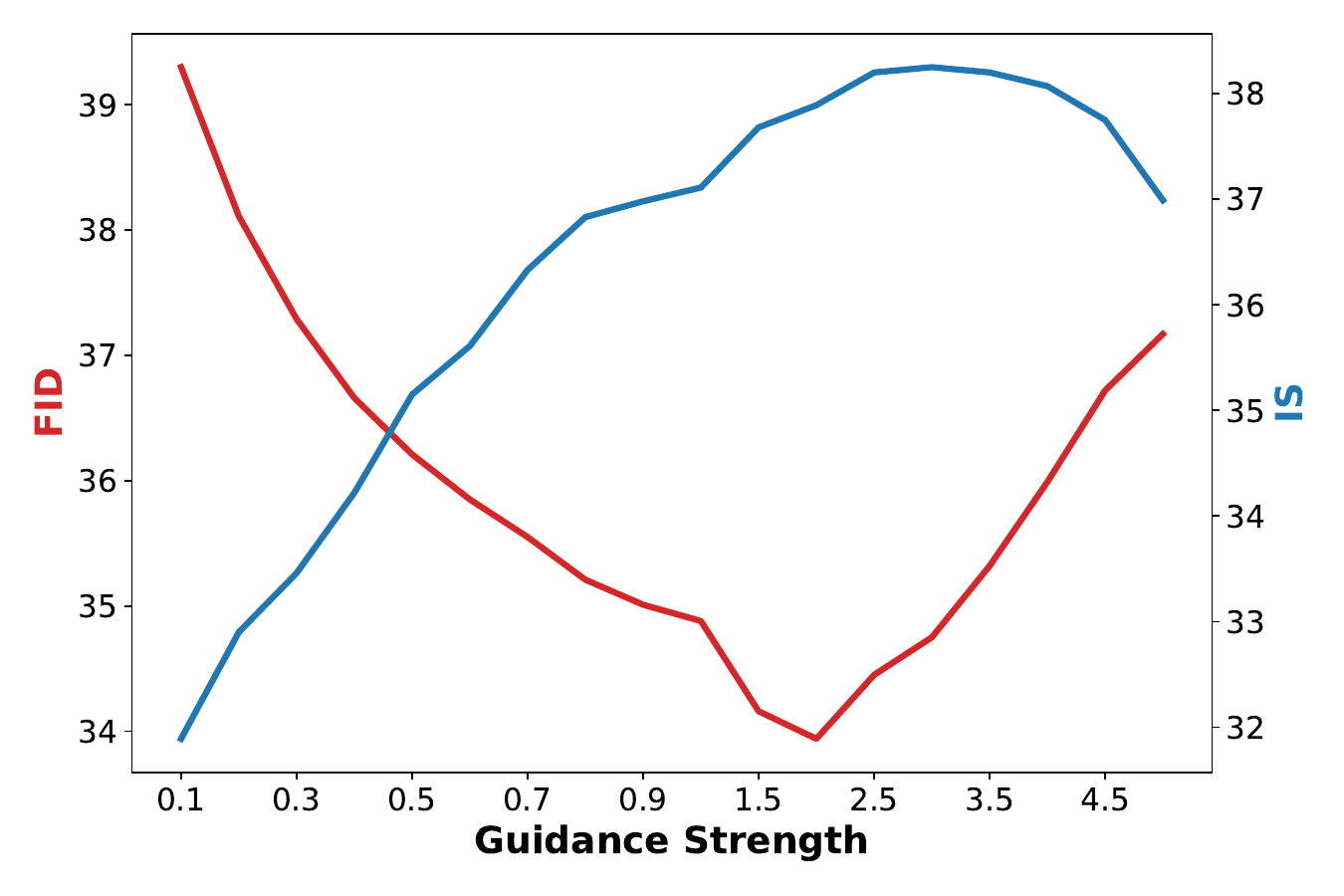}
    \caption{{The influence of guidance strength for offline guidance on ImageNet100 dataset.}}
    \label{fig:fid_vs_guidance_strength}
\end{figure}

\paragraph{Guidance strength \textit{v.s.} FID.} Figure~\ref{fig:fid_vs_guidance_strength} illustrates the relationship between FID and guidance strength, revealing an optimal point for achieving the highest fidelity in FID.

\begin{figure}
    \centering
\includegraphics[width=0.7\textwidth]{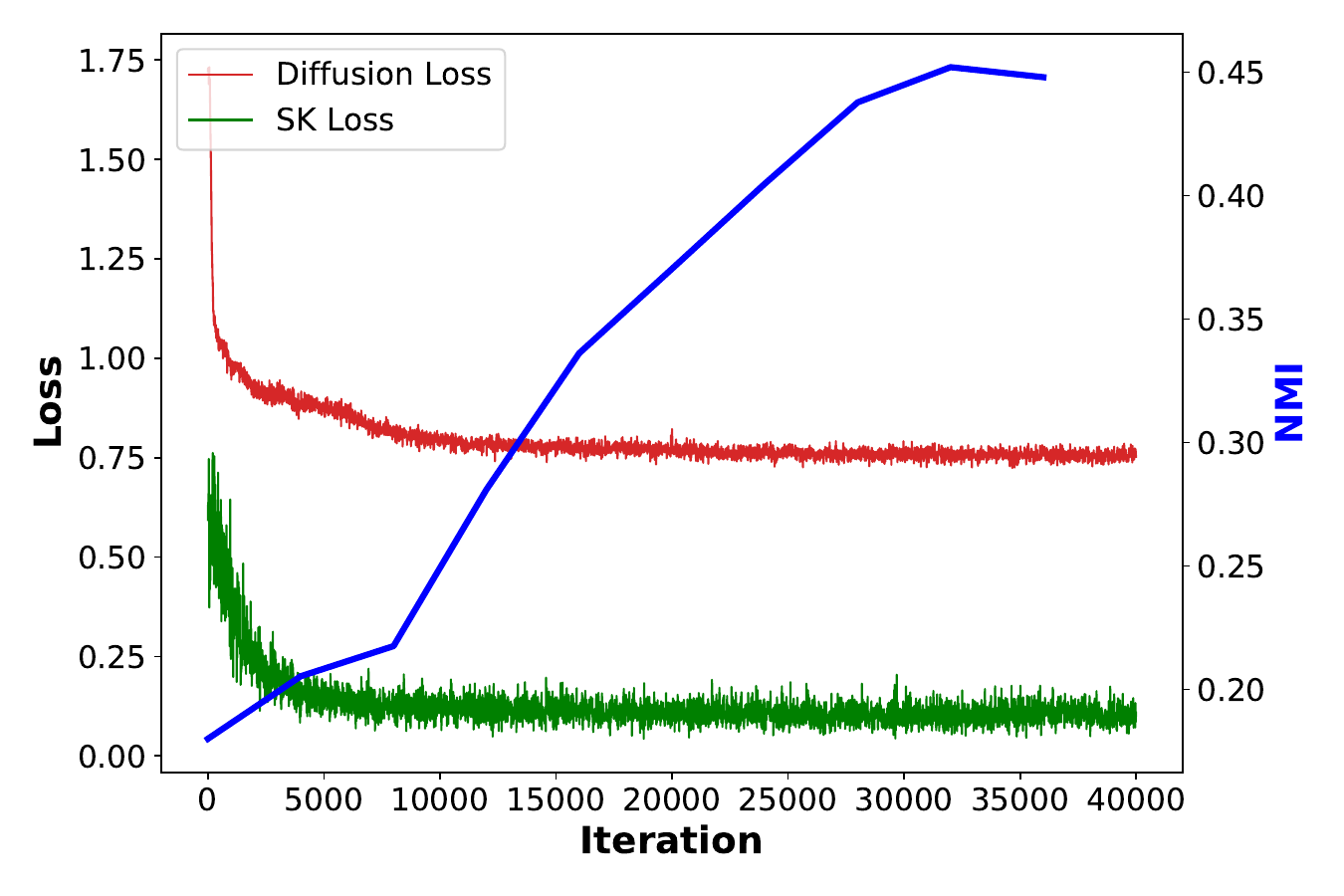}
    \caption{{The trend of Normalized Mutual Information (NMI) with respect to different loss functions during training.}}
\label{fig:loss_curve}
\end{figure}

\paragraph{Loss curve and NMI.} In Figure~\ref{fig:loss_curve}, we display the loss and NMI trends during training. Our loss shows stable convergence, and NMI steadily rises throughout training. This indicates the increasing discriminative power of the evolving features.  %

\section{Visualization Results}
\label{supp:more_vis}

\begin{figure}
\begin{subfigure}{.99\textwidth}
  \centering
  \includegraphics[width=.98\linewidth]{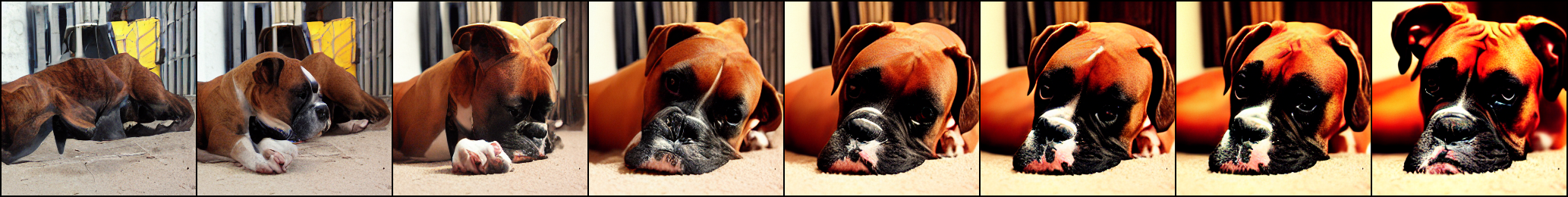}  
  \label{fig:sub-second}
\end{subfigure}
\newline
\begin{subfigure}{.99\textwidth}
  \centering
  \includegraphics[width=.98\linewidth]{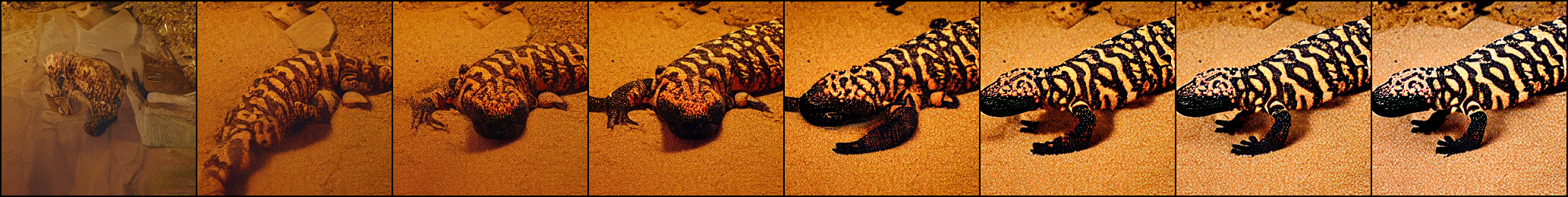}  
  \label{fig:sub-second}
\end{subfigure}
\newline
\begin{subfigure}{.99\textwidth}
  \centering
  \includegraphics[width=.98\linewidth]{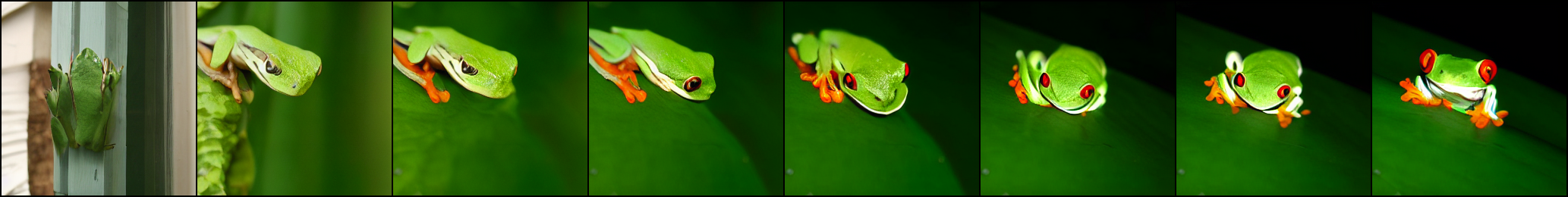}  
  \label{fig:sub-second}
\end{subfigure}
\newline
\begin{subfigure}{.99\textwidth}
  \centering
  \includegraphics[width=.98\linewidth]{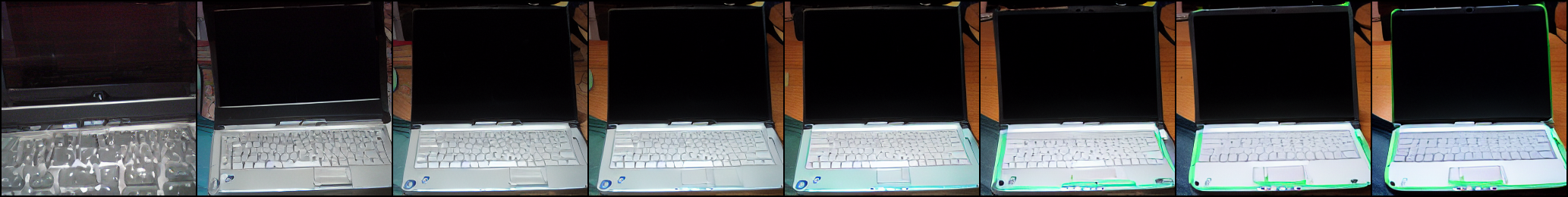}  
  \label{fig:sub-second}
\end{subfigure}
\newline
\begin{subfigure}{0.99\textwidth}
  \centering
  \includegraphics[width=.98\linewidth]{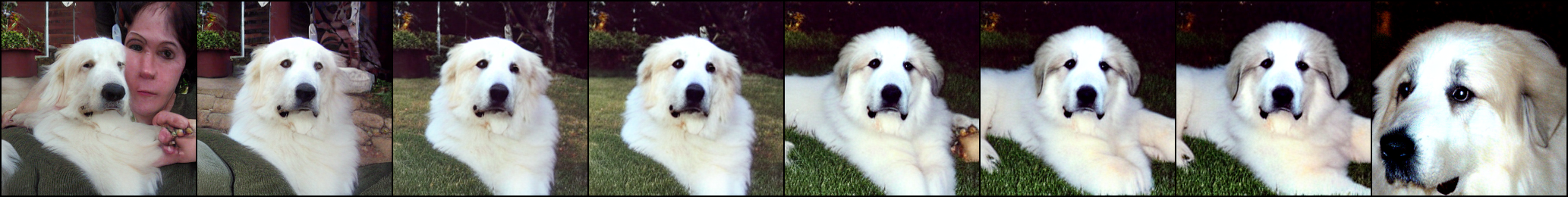}  
  \label{fig:sub-first}
\end{subfigure}
\newline
\caption{The visualization of incrementally increasing the guidance strength from 0 to 0.4 in flow matching on self-guidance within ImageNet256.}
\label{fig:vis_gs_in256}
\end{figure}

\begin{figure}
\begin{subfigure}{.99\textwidth}
  \centering
  \includegraphics[width=.98\linewidth]{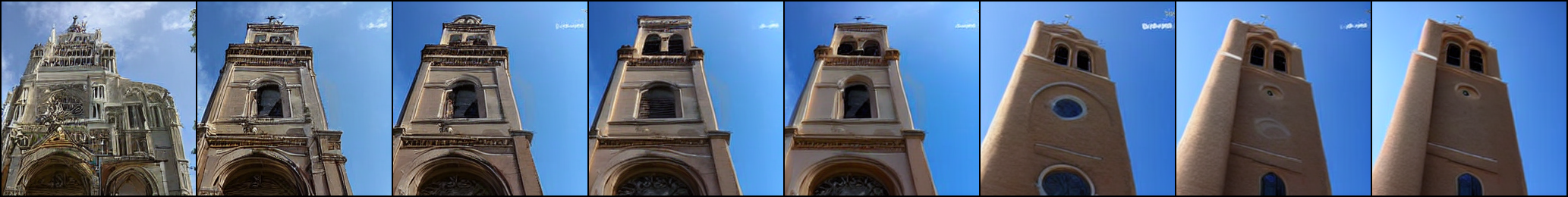}  
  \label{fig:sub-second}
\end{subfigure}
\newline
\begin{subfigure}{.99\textwidth}
  \centering
  \includegraphics[width=.98\linewidth]{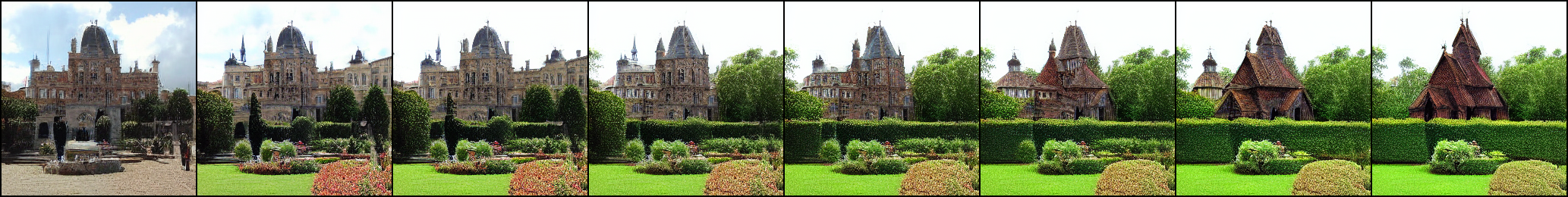}  
  \label{fig:sub-second}
\end{subfigure}
\newline
\begin{subfigure}{.99\textwidth}
  \centering
  \includegraphics[width=.98\linewidth]{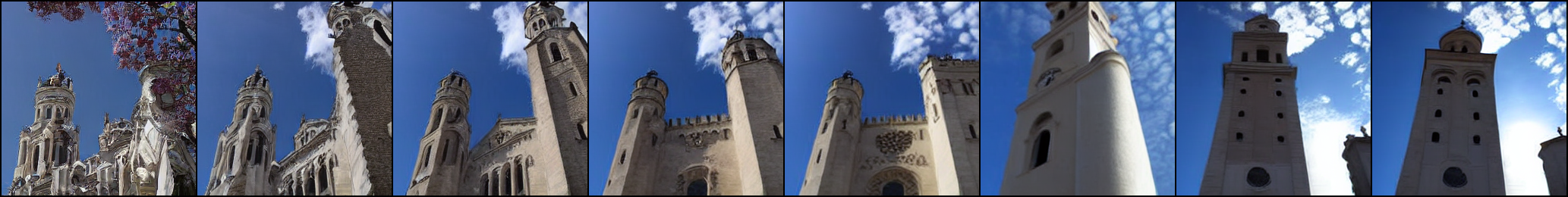}  
  \label{fig:sub-second}
\end{subfigure}
\caption{The visualization of incrementally increasing the guidance strength from 0 to 0.5 in flow matching on self-guidance within Churches256.}
\label{fig:vis_gs_churches}
\end{figure}

\paragraph{Visualization of various guidance strength.}  We visually showcase samples across different guidance strengths in Figure~\ref{fig:vis_gs_churches} and~\ref{fig:vis_gs_in256}. These illustrations illustrate that as the guidance strength ($g$) increases, the generated images exhibit higher levels of realism.

\begin{figure}
    \centering
    \includegraphics[width=.87\textwidth]
    {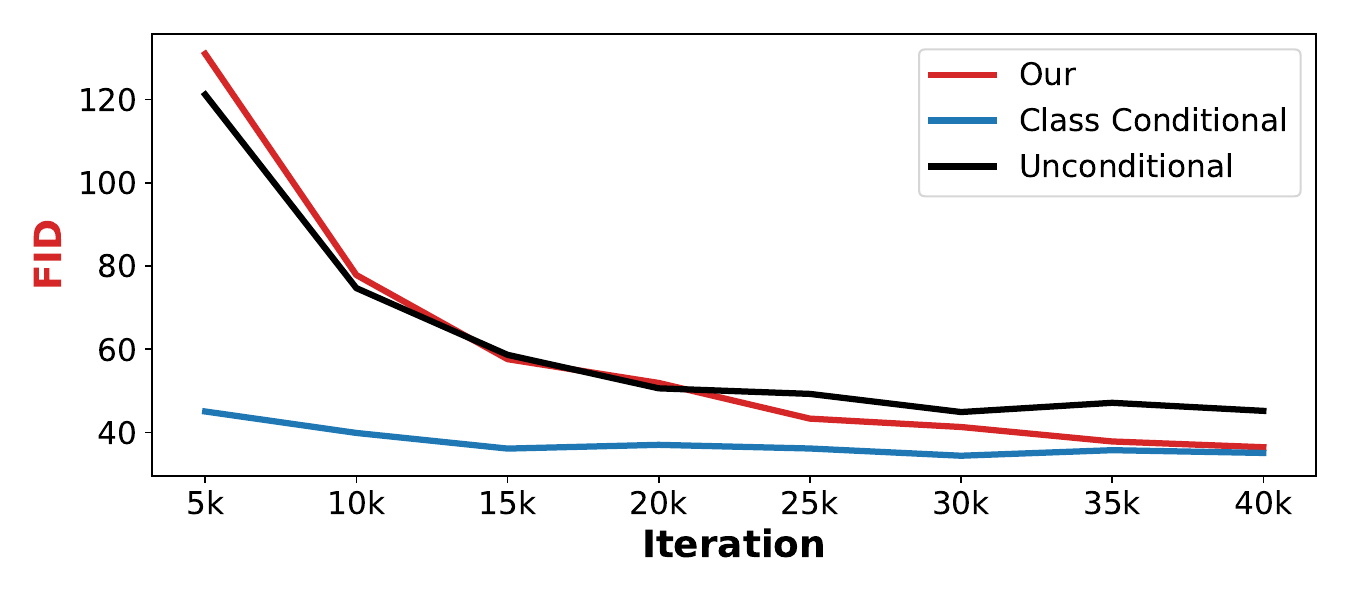}
    \vspace{-10pt}
    \caption{{Trend in FID  compared to unconditional diffusion models.}}
    \label{fig:fid_trend_comparison}
\end{figure}

\paragraph{FID trend.} In Figure~\ref{fig:fid_trend_comparison}, we compare the FID trends with unconditional and class-conditioned diffusion models. Initially, our method converges at a rate similar to that of unconditional diffusion models, but in the later stages, it converges more rapidly, achieving comparable levels to class-conditional diffusion models.

\begin{figure}
    \centering
    \includegraphics[width=.87\textwidth]{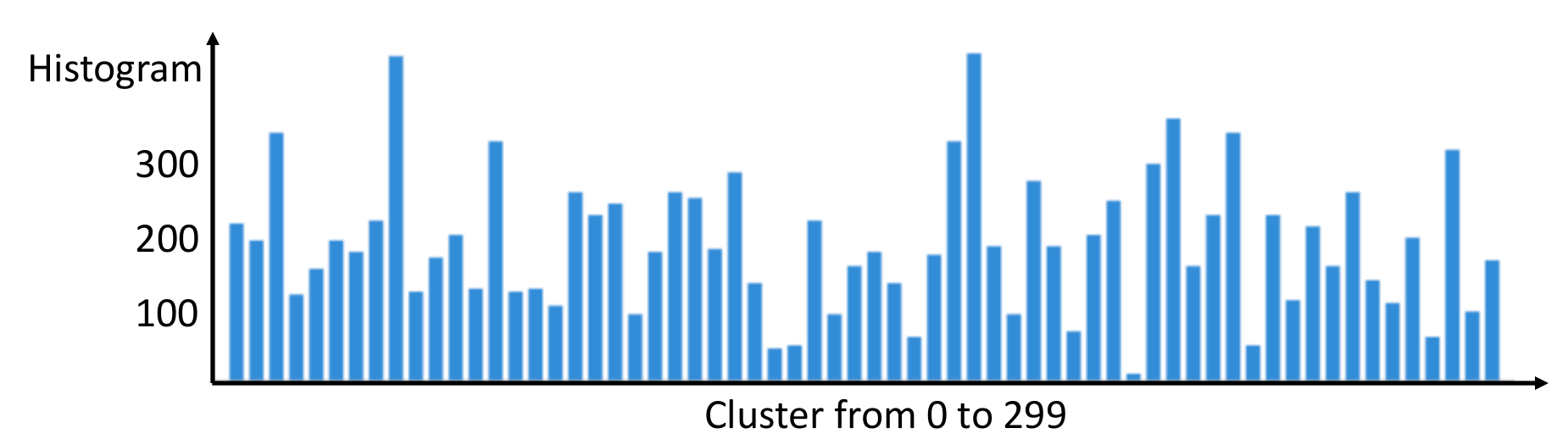}
    \caption{Histogram of cluster assignments in ImageNet100 with 300 clusters.}
    \label{fig:historgram_cluster_assignment}
\end{figure}

\paragraph{Clustering non-collapse Assignment.} To further analyze the clustering of our learned prototypes, we summarize the histogram of cluster assignments in Figure~\ref{fig:historgram_cluster_assignment}, providing evidence of non-collapse assignment during our training.

\begin{figure}
    \centering
    \includegraphics[width=.77\textwidth]{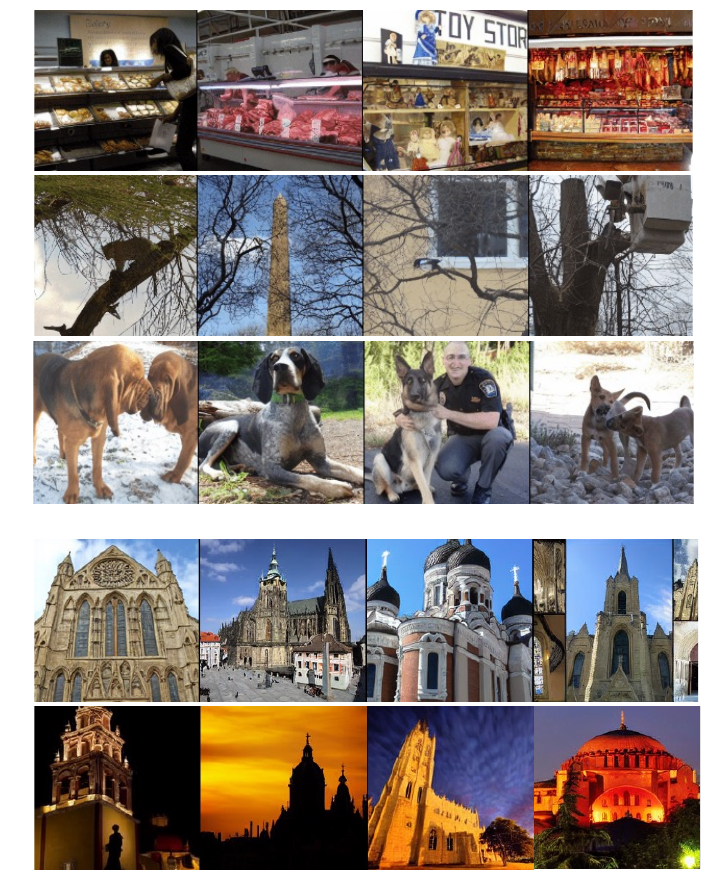}
    \caption{Cluster visualization of ImageNet256 and Churches.}
    \label{fig:cluster_vis}
\end{figure}

\paragraph{Cluster Visualization.} We provide visualizations of the offline clustering results in Figure~\ref{fig:cluster_vis}, showcasing the discriminative nature of features from diffusion models and the similarity in color and species of the generated images.

\begin{figure}
\begin{subfigure}{.99\textwidth}
  \centering
  \includegraphics[width=.98\linewidth]{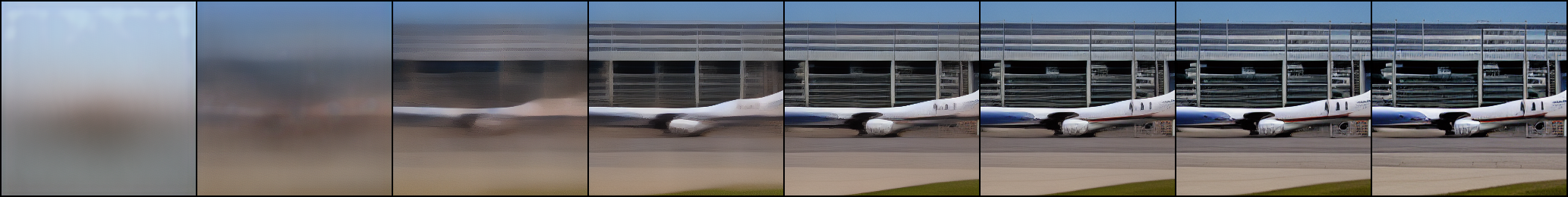}  
  \label{fig:sub-second}
\end{subfigure}
\newline
\begin{subfigure}{.99\textwidth}
  \centering
  \includegraphics[width=.98\linewidth]{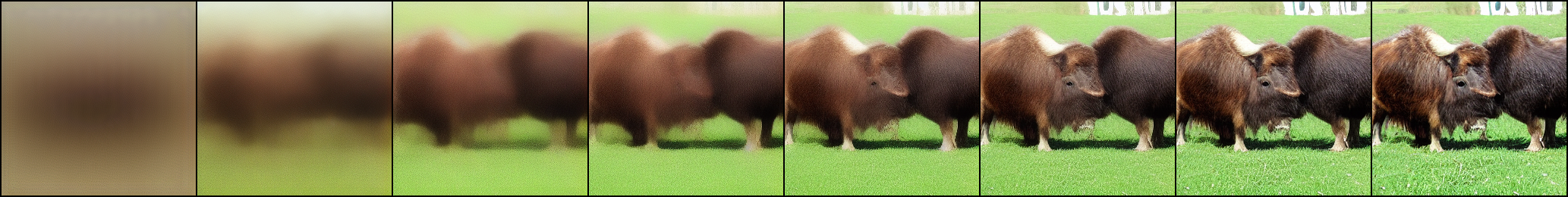}  
  \label{fig:sub-second}
\end{subfigure}
\newline
\begin{subfigure}{.99\textwidth}
  \centering
  \includegraphics[width=.98\linewidth]{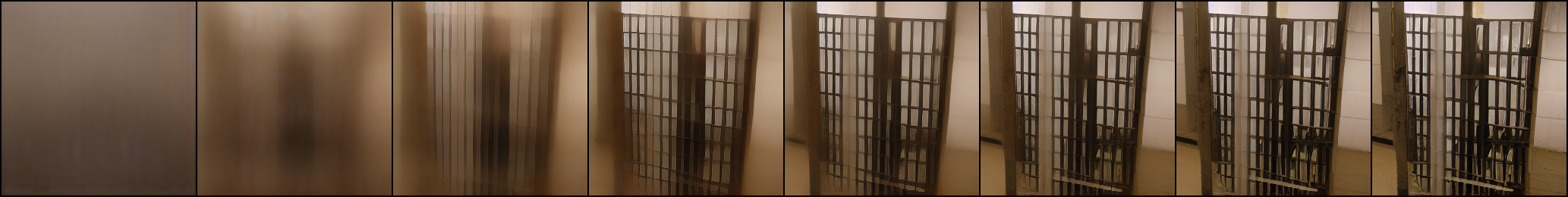}  
  \label{fig:sub-second}
\end{subfigure}
\newline
\begin{subfigure}{.99\textwidth}
  \centering
  \includegraphics[width=.98\linewidth]{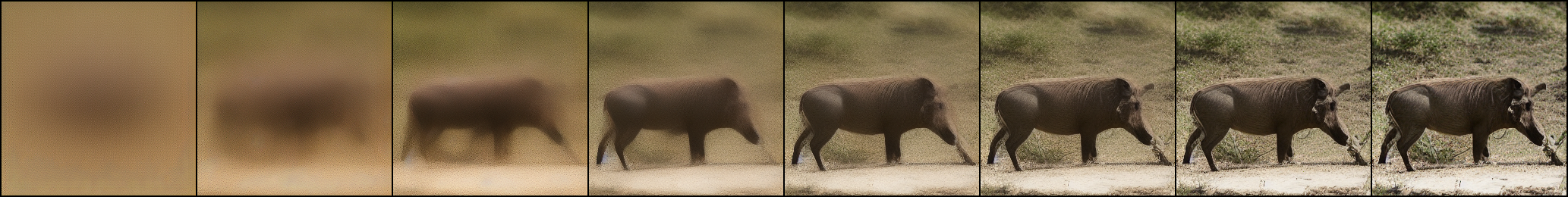}  
  \label{fig:sub-second}
\end{subfigure}
\newline
\caption{The visualization of  the estimation of $\bx_1$ on ImageNet256 during sampling with a guidance strength set to 0.4.}
\label{fig:flow_chains_in256}
\end{figure}

\begin{figure}
\begin{subfigure}{.99\textwidth}
  \centering
  \includegraphics[width=.98\linewidth]{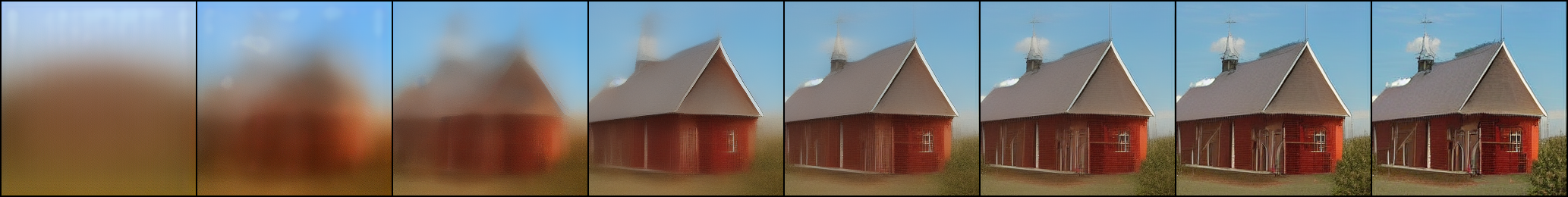}  
  \label{fig:sub-second}
\end{subfigure}
\newline
\begin{subfigure}{.99\textwidth}
  \centering
  \includegraphics[width=.98\linewidth]{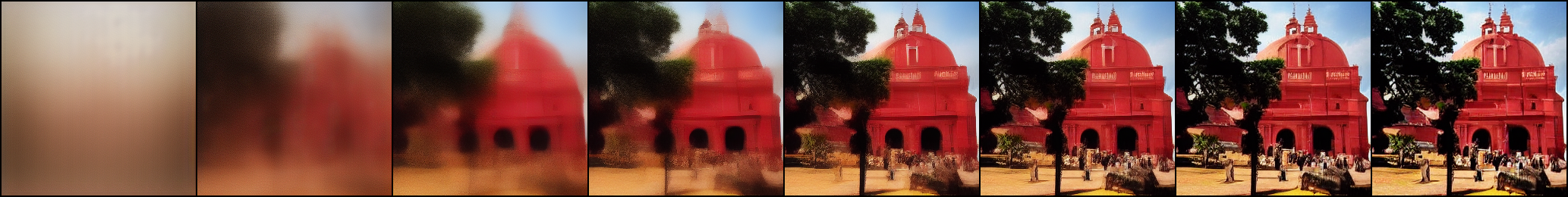}  
  \label{fig:sub-second}
\end{subfigure}
\newline
\begin{subfigure}{.99\textwidth}
  \centering
  \includegraphics[width=.98\linewidth]{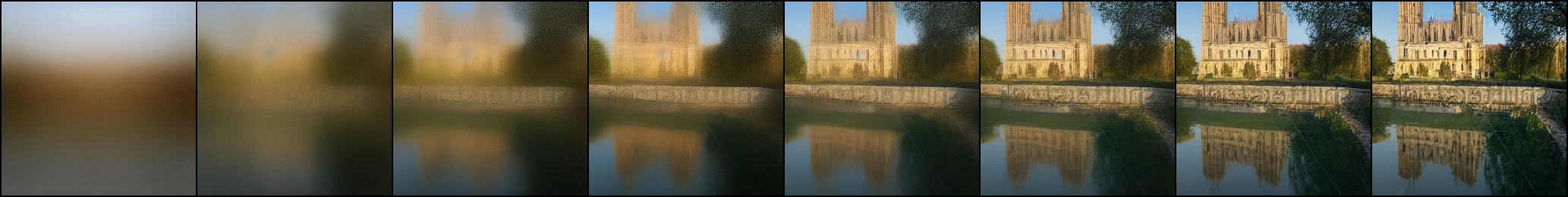}  
  \label{fig:sub-second}
\end{subfigure}
\newline
\begin{subfigure}{.99\textwidth}
  \centering
  \includegraphics[width=.98\linewidth]{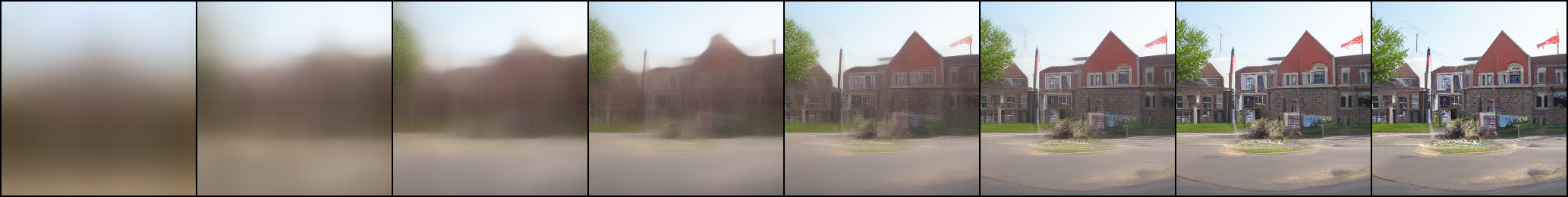}  
  \label{fig:sub-second}
\end{subfigure}
\newline
\caption{The visualization of the estimation of $\bx_1$ on Churches256 during sampling with a guidance strength set to 0.4.}
\label{fig:flow_chains_churches256}
\end{figure}

\paragraph{Diffusion Chains Visualization.} We illustrate the estimation of $\vx_1$ during sampling in Figure~\ref{fig:flow_chains_in256} (for ImageNet256) and in Figure~\ref{fig:flow_chains_churches256} (for LSUN-Churches256).

\begin{figure}
\begin{subfigure}{.99\textwidth}
  \centering
  \includegraphics[width=.98\linewidth]{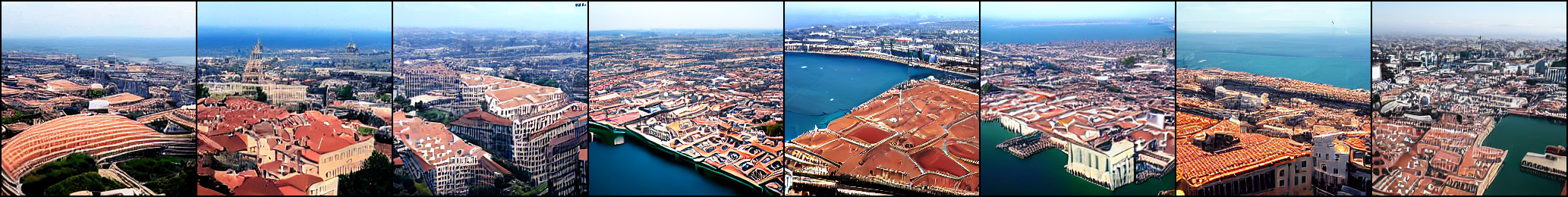}  
  \label{fig:sub-second}
\end{subfigure}
\newline
\begin{subfigure}{.99\textwidth}
  \centering
  \includegraphics[width=.98\linewidth]{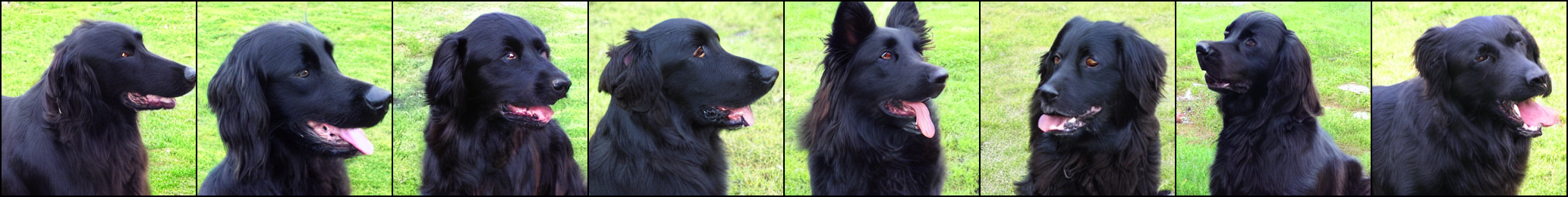}  
  \label{fig:sub-second}
\end{subfigure}
\newline
\begin{subfigure}{.99\textwidth}
  \centering
  \includegraphics[width=.98\linewidth]{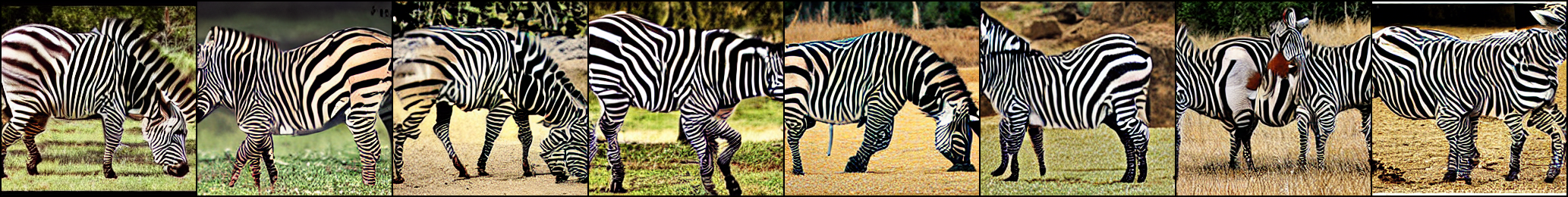}  
  \label{fig:sub-second}
\end{subfigure}
\newline
\begin{subfigure}{.99\textwidth}
  \centering
  \includegraphics[width=.98\linewidth]{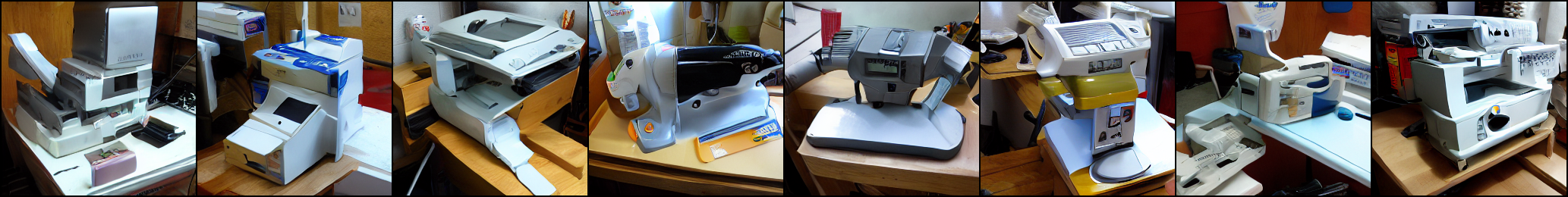}  
  \label{fig:sub-second}
\end{subfigure}
\newline
\caption{Offline guidance sampling visualization from different cluster ids.}
\end{figure}

\end{document}